\documentclass{article}

    \usepackage[preprint]{neurips_2025}

\usepackage[utf8]{inputenc} % allow utf-8 input
\usepackage[T1]{fontenc}    % use 8-bit T1 fonts
\usepackage{url}            % simple URL typesetting
\usepackage{booktabs}       % professional-quality tables
\usepackage{amsfonts}       % blackboard math symbols
\usepackage{nicefrac}       % compact symbols for 1/2, etc.
\usepackage{microtype}      % microtypography

\usepackage{thmtools}
\usepackage{thm-restate}
\usepackage{hyperref}       % hyperlinks

\newtheorem{theorem}{Theorem}
\newtheorem{example}{Example}
\newtheorem{definition}[theorem]{Definition}
\newtheorem{lemma}[theorem]{Lemma}
\newtheorem{proposition}[theorem]{Proposition}
\newtheorem{corollary}[theorem]{Corollary}
\newtheorem{remark}[theorem]{Remark}

\usepackage[utf8]{inputenc} % allow utf-8 input
\usepackage[T1]{fontenc}    % use 8-bit T1 fonts
\usepackage[dvipsnames]{xcolor}
\usepackage{hyperref}       % hyperlinks
\usepackage{url}            % simple URL typesetting
\usepackage{booktabs}       % professional-quality tables
\usepackage{amsfonts,bbm}       % blackboard math symbols
\usepackage{nicefrac}       % compact symbols for 1/2, etc.
\usepackage{microtype}      % microtypography
\usepackage{amsmath}
\usepackage{amssymb}
\usepackage{thmtools}
\usepackage{tikz}
\usetikzlibrary{decorations.pathreplacing, arrows.meta, positioning}

\DeclareMathOperator{\CS}{CS}
\DeclareMathOperator{\SD}{SD}
\DeclareMathOperator{\prox}{prox}
\DeclareMathOperator{\VS}{VS}
\DeclareMathOperator{\MQ}{MQ}
\DeclareMathOperator{\DL}{DL}
\DeclareMathOperator{\pmon}{mmon}

\DeclareMathOperator{\parity}{par}

\let\vec\mathbf

\newcommand{\VCD}{\mathrm{VCD}}
\newcommand{\EX}{\mathrm{EX}}

\newcommand{\cX}{{\mathcal X}}
\newcommand{\cC}{{\mathcal C}}

\newcommand{\cS}{{\mathcal S}}
\newcommand{\cL}{{\mathcal L}}

\newcommand{\seq}{\subseteq}

\newcommand{\sm}{\setminus}
\newcommand{\reals}{\mathbbm{R}}
\newcommand{\CLAUSES}{\mathcal{C}_{\operatorname{claus}}}
\newcommand{\MONOMIALS}{\mathcal{C}_{\operatorname{mon}}}
\newcommand{\BOXES}{\mathcal{C}_{\operatorname{box}}}
\newcommand{\MDNF}[3]{\mathcal{C}_{\operatorname{MDNF}}^{#1, #2, #3}}

\newcommand{\ol}[1]{\overline{#1}}
\usepackage{xspace}

\newif\iffinal
\iffinal
    \newcommand{\YC}[1]{}
\else
    \newcommand{\YC}[1]{\todo[fancyline,color=Maroon!40]{YC: #1}\xspace}
\fi

\title{Formal Models of Active Learning from Contrastive Examples}

\author{%
  Farnam Mansouri \\
 University of Waterloo \\
\texttt{f5mansou@uwaterloo.ca} 
\And
Hans U. Simon \\
MPI-INF and Ruhr-University Bochum\\
    \texttt{hsimon@mpi-inf.mpg.de} \\
\AND
  Adish Singla \\
  MPI-SWS\\
    \texttt{adishs@mpi-sws.org} \\
  \And
  Yuxin Chen\\
  University of Chicago \\
  \texttt{chenyuxin@uchicago.edu} \\
  \And
  Sandra Zilles \\
  University of Regina \\
\texttt{zilles@cs.uregina.ca} \\
}
\begin{document}

\maketitle

\begin{abstract}
  Machine learning can greatly benefit from providing learning algorithms with pairs of contrastive training examples---typically pairs of instances that differ only slightly, yet have different class labels. Intuitively, the difference in the instances helps explain the difference in the class labels. This paper proposes a theoretical framework in which the effect of various types of contrastive examples on active learners is studied formally. The focus is on the sample complexity of learning concept classes and how it is influenced by the choice of contrastive examples. We illustrate our results with geometric concept classes and classes of Boolean functions. Interestingly, we reveal a connection between learning from contrastive examples and the classical model of self-directed learning.
\end{abstract}

\section{Introduction}\label{sec:intro}

In machine learning, contrastive data has been used for various purposes, most notably in order to reduce the number of training samples needed to make high-quality predictions, or to explain the predictions of black-box models.
A contrastive example for a labeled training data point $(x,b)$ with label $b\in\{0,1\}$ could be, e.g., a labeled point
$(x',1-b)$, such that $x'$ fulfills some additional predefined property. This additional property
could be defined in a way so as to convey helpful information for learning. 
For instance, a constraint on
$x'$ could be that it is the \emph{closest}\/ point to $x$ (in the given pool of data points, and with respect to a fixed
underlying metric) that has a different label from $b$. Such examples are often called counterfactuals %in the literature 
and have proven useful in various learning settings, including explanation-based supervised learning \cite{DasguptaDRS18}, but also reinforcement learning \cite{HuberDMOA23,OlsonKNLW21}, and learning recommender systems \cite{SaitoJ21}.

Intuitively, counterfactuals present a learning algorithm with an important feature or property of a data
object that crucially affects its classification. For example, a training data point $x$ might be an image of a
cat, together with its class label (the subspecies of cat). A counterfactual piece of information might be
an image $x'$ of a cat that looks similar to the cat in $x$, yet belongs to a different subspecies. The main
feature distinguishing $x'$ from $x$ could be highlighted in the image, as a means of explaining the
classification of $x$. Similarly, contrastive information can be used in natural language processing and many
other applications, and was applied successfully in representation learning, see, for example, \cite{AshGKM22,ChenK0H20,LiangZY20,MaZMS21}. Interestingly, also human learners exhibit improved classification
abilities when they are taught with the help of such visual explanations \cite{MacAodhaSCPY18}.

Theoretical studies %on contrastive learning %, %which are closer to our work, 
%define and 
have analyzed
the sample complexity in specific models of learning with contrastive examples, such as PAC-learning of
distance functions \cite{alon23}, or active learning with a greedy teacher \cite{WangSC21}, %or active learning with feature feedback from 
a randomized teacher \cite{PoulisD17} or an adversarial teacher \cite{DasguptaS20}. 
Other formal studies focus on specific aspects of learning with contrastive information, for instance, how
negative examples influence the learning process \cite{AshGKM22} or how individual loss functions affect
contrastive learning \cite{HaoChenWGM21}.

%Contrastive information in general is of relevance in many contexts of learning or synthesis through interaction. %For example, while most research on contrastive learning focusses on supervised settings (in
%which data is presented in the form of (datum, label) pairs), research on reinforcement learning has also
%explored the benefits of counterfactual information \cite{HuberDMOA23,OlsonKNLW21}. In a similar vein, research on
%recommender systems deploys recommendation policies built on counterfactual estimators, which
%estimate the performance of alternate recommendation policies on the data acquired with the current
%policy \cite{SaitoJ21}. 
One example of the use of contrastive information, particularly relevant to our study, is counter-example-guided program
synthesis (CEGIS) in formal methods research \cite{AbateDKKP18}. Consider the task of synthesizing a program that satisfies a given specification $S$. In CEGIS, a synthesizer generates a candidate program $P$, which is then checked against the specification $S$ by a verifier. If $P$ violates $S$, the verifier returns a counterexample witnessing the violation. This process is iterated until a program satisfying $S$ is obtained. We can see each candidate program $P$ as a query to the verifier, which acts as an oracle. Importantly, in this setting, the oracle follows a \emph{known procedure}. In particular, one can implement the oracle to return a specific counterexample (e.g., a smallest one) that acts as a contrastive example, and the synthesizer can be \emph{tailored to the knowledge of how the counterexample is generated}. For example, if the synthesizer knows that a smallest counterexample is generated, it can rule out all programs for which there would have been smaller counterexamples.

Motivated by such applications in program synthesis, the main goal of this paper is to introduce and analyze formal settings of learning from contrastive examples, specifically modeling situations in which the learner has \emph{knowledge of the way in which the oracle selects contrastive examples}. 

We propose a generic formal framework, encompassing various ways in which
contrastive examples are chosen (i.e., not just as counterfactuals); this is the first contribution of our paper. We then analyze several settings in this framework w.r.t.\ the resulting sample complexity, i.e., the number of examples required for identifying the underlying target concept, resulting in various non-trivial upper and lower bounds on learning classes of geometric objects or Boolean functions (most notably 1-decision lists and monotone DNFs); this constitutes our second main contribution. Finally, as a third contribution, we reveal a surprising connection between the sample complexity %in one of our models 
of contrastive learning and the mistake bound in the classical notion of self-directed (online) learning \cite{GoldmanS94}.

%\section{Preliminaries}
%\label{sec:preliminaries}

\section{Models of Contrastive Learning}
\label{sec:models}

Let $\mathcal{X}$ be any (finite or infinite) domain that is bounded under some  metric. A concept over $\mathcal{X}$ is a subset of $\mathcal{X}$; we will identify a concept $C\subseteq\mathcal{X}$ with its indicator function on $\mathcal{X}$, given by $C(x)=1$ if $x\in C$, and $C(x)=0$ if $x\in\mathcal{X}\setminus C$. A concept class over $\mathcal{X}$ is a set of concepts over $\mathcal{X}$.
If $C$ and $C'$ are two concepts over $\mathcal{X}$, then $\Delta(C,C')$ refers to the symmetric difference of $C$ and $C'$. Moreover, $\delta(C,C')$ is the relative size of $\Delta(C,C')$ over $\mathcal{X}$ measured under the standard uniform distribution. In case $\mathcal{X}$ is finite, this is just $|\Delta(C,C')|/|\mathcal{X}|$.

$\mathcal{S}[\MQ](\mathcal{C})$ denotes the minimum worst-case number of membership queries \cite{Angluin88} for learning concepts in $\cC$, where the worst case is taken over all concepts in $\mathcal{C}$ and the minimum is taken over all membership query learners. Moreover, $\VCD(\cC)$ denotes the VC dimension of $\cC$.

Suppose a concept class $\mathcal{C}$ is fixed and an active learner asks membership queries about an unknown target concept $c\in\mathcal{C}$. In addition to the label of the queried example, the oracle might provide a contrastive example, e.g., a similar example of opposite label. We propose a more generic setting in which %a passive learner (receiving labeled examples from some underlying distribution) or 
the active learner is aided by an oracle that provides labeled examples complementing the %randomly received or 
actively selected ones in \emph{some} pre-defined form (using a so-called contrast set). The learner tries to identify an unknown target concept $C^*\in\mathcal{C}$ in sequential rounds. In round $1$, the version space (i.e., the set of possible target concepts under consideration) is $\mathcal{C}_1:=\mathcal{C}$. Round $i$, $i\ge 1$, is as follows:

\begin{enumerate}
    \item The learner %receives a randomly sampled instance $x_i\in\mathcal{X}$ (passive case) or 
    poses a membership query consisting of an instance $x_i\in\mathcal{X}$.
    \item The oracle's response is a pair $[y_i,(x_i',y_i')]\in\{0,1\}\times(\cX\times\{0,1\})$ such that:
    \begin{itemize}
    \item $y_i=C^*(x_i)$ and $y'_i=C^*(x'_i)$, i.e., $y_i$ and $y'_i$ are the correct labels of $x_i$ and $x'_i$, resp., under the target concept,
    \item $x_i'\in\CS(x_i,C^*,\mathcal{C}_i)$, where $\CS(x_i,C^*,\mathcal{C}_i)\subseteq\mathcal{X}$ is a set of instances  called the \emph{contrast set}\/ of $x_i$ wrt $C^*$ and $\mathcal{C}_i$. The oracle is adversarial, in that it can pick any $x'_i\in\CS(x_i,C^*,\mathcal{C}_i)$ as part of the response to query $x_i$. If $\CS(x,C^*,\mathcal{C}_i)$ is empty, the oracle returns a dummy response $\omega$, to convey that no contrastive example exists.
\end{itemize}
%While technically the oracle's response is a tuple with multiple components, 
%(We often speak about the oracle's response while referring to just one part of that response (e.g., we may say the oracle returns the contrastive example $(x'_i,C^*(x'_i))$).)
\item The version space is updated; in particular, the new version space $\mathcal{C}_{i+1}$ consists of all concepts $C$ in $\mathcal{C}_i$ for which $C(x_i)=y_i$ and:
\begin{itemize}
\item $\CS(x_i,C,\mathcal{C}_i)=\emptyset$ in case the dummy response $\omega$ is received,
    \item $C(x'_i)=y'_i$ and $x'_i\in\CS(x_i,C,\mathcal{C}_i)$ otherwise.
\end{itemize}
%We write $\mathcal{C}_{i+1}=\update_{\CS}(\mathcal{C}_i,x_i,y_i,x'_i,y'_i)$ for short.
\end{enumerate}

This definition assumes that the learner knows the function mapping any tuple $(x_i,C,\mathcal{C}_i)$ %\in\mathcal{X}\times\mathcal{C}\times 2^{2^\cX}$ 
to the set $\CS(x_i,C,\mathcal{C}_i)$.
%Learning proceeds sequentially; in round $i$, the learner will issue query $x_i\in\mathcal{X}$ and receive response $[C^*(x_i),(x'_i,C^*(x'_i))]\in\{0,1\}\times(\mathcal{X}\times\{0,1\})$,  assuming 
\iffalse
The specific queries and responses in the first $n$ rounds then form an \emph{interaction sequence}\/ of length $n$, given by
\[[(x_1,y_1),(x'_1,y'_1)],\ldots,[(x_n,y_n),(x'_n,y'_n)]\,.\] 
In particular, the \emph{version space}\/ of the learner after this interaction sequence consists of all concepts $C\in\mathcal{C}$ with the following three properties:
\begin{itemize}
    \item $C(x_i)=y_i$ for $1\le i\le n$,
    \item $C(x'_i)=y'_i$ for $1\le i\le n$,
    \item $x'_i\in\CS(x_i,C)$ for $1\le i\le n$.
\end{itemize}
We denote this version space by $\VS([(x_1,y_1),(x'_1,y'_1)],\ldots,[(x_n,y_n),(x'_n,y'_n)])$. 
\fi
The core difference to the traditional notion of version space is that the learner can reduce the version space by knowing the contrast set mapping $\CS$: a concept $C$, even if consistent with all seen labeled examples, is excluded from the version space if $x'_i\notin\CS(x_i,C,\mathcal{C}_i)$ for some $i$.

%Suppose $(x_1,\ldots,x_n)$ is a sequence of queries, and $[y_1,(x'_1,y'_1)],\ldots,[y_n,(x'_n,y'_n)]$ the sequence of the corresponding answers by the oracle. 
The labeled examples collected in 
the first $n$ rounds, together with the corresponding version spaces, then form an \emph{interaction sequence}\/ of length $n$ wrt $\CS$, given by
\[[(x_1,y_1),(x'_1,y'_1),\mathcal{C}_2],\ldots,[(x_n,y_n),(x'_n,y'_n),\mathcal{C}_{n+1}]\,.\] 
This interaction sequence \emph{$\varepsilon$-approximates}\/ the target concept $C^*\in\mathcal{C}$ (for some $\varepsilon\ge 0$) iff $\mathcal{C}_{n+1}\subseteq\{C\in\mathcal{C}\mid \delta(C^*,C)\le\varepsilon\}$. 
The sequence \emph{identifies}\/ the target concept $C^*\in\mathcal{C}$ iff $\mathcal{C}_{n+1}=\{C^*\}$.

Accordingly, we define the following notion of sample complexity.

\begin{definition}
Assume $\mathcal{X}$, $\mathcal{C}$, $\CS$, and $\varepsilon$ are fixed. 
%A learner is a mapping $L$ that takes as input a (possibly empty) sequence $[(x_1,y_1),(x'_1,y'_1)],\ldots,[(x_i,y_i),(x'_i,y'_i)]$, and outputs a datum $x_{i+1}\in\mathcal{X}$. $L$ is said to produce $\sigma=[(x_1,y_1),(x'_1,y'_1),\mathcal{C}_2],\ldots,[(x_n,y_n),(x'_n,y'_n),\mathcal{C}_{n+1}]$ iff (i) $L$ outputs $x_1$ on input of the empty sequence and, for all $i$ such that $1\le i<n$, the output of $L$ on input $[(x_1,y_1),(x'_1,y'_1)],\ldots,[(x_i,y_i),(x'_i,y'_i)]$ is $x_{i+1}$, and (ii) $\sigma$ is a valid interaction sequence wrt $\CS$.
%\item The passive contrast sample complexity on $C^*$ wrt $\mathcal{C}$, denoted by $S^p_{\CS}(C^*,\mathcal{C},\varepsilon)$, is the \textcolor{red}{(expected ???)} length of any interaction sequence that $\varepsilon$-approximates $C^*\in\mathcal{C}$. The passive contrast sample complexity of $\mathcal{C}$, denoted by $S^p_{\CS}(L,\mathcal{C})$, is then defined as $S^p_{\CS}(\mathcal{C},\varepsilon)=\sup_{C^*\in\mathcal{C}}S^p_{\CS}(C^*,\mathcal{C},\varepsilon)$.
The $\varepsilon$-approximate contrast sample complexity of a learner $L$ on $C^*$ wrt $\mathcal{C}$, denoted by $\cS_{\CS}(L,C^*,\mathcal{C},\varepsilon)$, is the largest length of any valid interaction sequence $\sigma=[(x_1,y_1),(x'_1,y'_1),\mathcal{C}_2],\ldots,[(x_n,y_n),(x'_n,y'_n),\mathcal{C}_{n+1}]$ that $\varepsilon$-approximates $C^*\in\mathcal{C}$ and in which $L$ chooses $x_{i+1}$ on input of $[(x_1,y_1),(x'_1,y'_1)],\ldots,[(x_i,y_i),(x'_i,y'_i)]$, for all $i$. (Here the worst case length is taken over the sequence of possible choices of $x_i'$ made by the oracle.) Finally, $\cS_{\CS}(L,\mathcal{C},\varepsilon)=\sup_{C^*\in\cC}\cS_{\CS}(L,C^*,\cC,\varepsilon)$, and $\mathcal{S}_{\CS}(\mathcal{C},\varepsilon)=\inf_{L}\cS_{\CS}(L,\mathcal{C},\varepsilon)$.
For exact learning, we consider only interaction sequences that identify $C^*$, and write $\mathcal{S}_{\CS}(\mathcal{C})$ instead of $\mathcal{S}_{\CS}(\mathcal{C},\varepsilon)$.
    %\item For any learner $L$ and any $C^*\in\mathcal{C}$, let the contrast sample complexity of $L$ on $C^*$ wrt $\mathcal{C}$, denoted by $S_{\CS}(L,C^*,\mathcal{C})$, be the largest length of any interaction sequence that identifies $C^*\in\mathcal{C}$ and is produced by $L$. (Here the worst case length is taken over the sequence of possible choices of $x_i'$ made by the oracle.)
    %\item For any learner $L$, let the contrast sample complexity of $L$ on  $\mathcal{C}$, denoted by $S_{\CS}(L,\mathcal{C})$, be the defined as $S_{\CS}(L,\mathcal{C})=\sup_{C^*\in\mathcal{C}}S_{\CS}(L,C^*,\mathcal{C})$.
    %\item The contrast sample complexity of $\mathcal{C}$, denoted by $\mathcal{S}_{\CS}(\mathcal{C})$, refers to the best-case contrast sample complexity taken over all learners $L$, i.e., $\mathcal{S}_{\CS}(\mathcal{C})=\inf_{L}S_{\CS}(L,\mathcal{C})$.
%\end{enumerate}
\end{definition}

Our model assumes the learner has perfect knowledge of the set $\CS$ of candidates from which contrastive examples are chosen. This assumption, while often too strong in practice, is indeed realistic in program synthesis settings, where the learning algorithm knows the rules by which the oracle selects counterexamples, as discussed in Section~\ref{sec:intro} for CEGIS \cite{AbateDKKP18}. Here the designer has full control over both the oracle and the learner. In addition, our model is useful for further reasons:
\begin{itemize}
    \item It can be  easily modified to address additional real-world settings, e.g., one can assume that the learner's notion of $\CS$ is not identical, but similar to that used by the oracle.
    \item Lower bounds from our strong model (e.g., our bound in terms of the self-directed learning complexity, Theorem~\ref{thm:DL-SD}) immediately transfer to such weakened versions of the model.
    %\item Formal methods research indeed does study settings in which the learning algorithm knows exactly the rules by which the oracle selects counterexamples, for instance in counterexample-guided synthesis \cite{AbateDKKP18}. Here the designer has full control over both the oracle and the learner.  
\end{itemize}

A first observation is that, without limiting the choice of the contrast set $\CS$, the contrast oracle is extremely powerful; any encoding of concepts in $\cC$ as subsets of $\cX$ of size at most $k$ can be used to define contrast sets witnessing a sample complexity of at most $k$:

\begin{restatable}{proposition}{propinjective}
Let $\mathcal{C}$ be a countable concept class over a countable $\mathcal{X}$. Let $T:\mathcal{C}\rightarrow 2^\mathcal{X}$ be any injective function that maps every concept in $\mathcal{C}$ to a finite set of instances. Then: 
\begin{enumerate}
\item There is some $\CS$ with $\mathcal{S}_{\CS}(\mathcal{C})\le\sup_{C\in\mathcal{C}}|T(C)|+1$. 
\item If $T(C)\not\subseteq T(C')$ for $C\ne C'$, then there is some $\CS$ such that $\mathcal{S}_{\CS}(\mathcal{C})\le\sup_{C\in\mathcal{C}}|T(C)|$.
\end{enumerate}
In particular, if $\cX$ is finite, then there is some $\CS$ such that $\mathcal{S}_{\CS}(\mathcal{C})\le 1+\min\{k\mid\sum_{i=0}^k\binom{|\cX|}{i}\ge|\cC|\}$. If $\mathcal{X}$ is countably infinite, then there is some $\CS$ such that $\mathcal{S}_{\CS}(\mathcal{C})=1$. 
\end{restatable}
\emph{Proof.} \emph{(Sketch.)} Let $x_1,x_2,\ldots$ be a repetition-free enumeration of  $\mathcal{X}$, and 
$\CS(x_i,C) = \{x_{j'}\}$ for $j' = \min\{j\mid j \ge i,\ x_j \in T(C)\}$.
A learner sets $n_1=1$ and starts with iteration 1. In iteration $i$, it asks a query for $x_{n_i}$. If it receives $x'_i=x_j$ as a contrastive example, then $j\ge n_i$. The learner will then set $n_{i+1}=j+1$ and proceed to iteration $i+1$. See the appendix for details.
\hfill$\Box$

Thus, unlimited choice of the mapping $\CS$ reduces learning from a contrast oracle to decoding a smallest encoding of concepts as sample sets. Therefore, the remainder of our study focuses on more natural choices of contrast sets. In particular, when fixing a function $d:\mathcal{X}\times\mathcal{X}\rightarrow\mathbb{R}^{\ge 0}$ (interpreted as a notion of distance), two natural choices of contrast set mappings are the following:
\paragraph{Minimum distance model.} Our first contrast set mapping, dubbed the \emph{minimum distance}\/ model, makes the oracle provide an example closest to $x$, among those that yield a different label under $C$ than $x$. For discrete $\mathcal{X}$, this translates to  $\CS^d_{\min}(x,C)=\arg\min\{d(x,x')\ |\ x' \in \cX,\ C(x') \neq C(x) \}$. %, i.e., the contrast set contains all instances closest to $x$, among those that yield a different label under $C^*$ than $x$. 
    For continuous $\mathcal{X}$, it can happen that no point  $x'\in \mathcal{X}$ with $C(x')\ne C(x)$ attains the infimum of the values $d(x,x')$ for such $x'$.\footnote{For instance, if $C$ is a closed interval in $\mathbb{R}$ and the query point $x$ lies inside this interval, then there is no point outside the interval that has distance closer to $x$ than any other point outside the interval.} To accommodate this case, we define $\CS^d_{\min}(x,C)=\arg\min\{d(x,x')\ |\ x' \in \cX,\ x'=\lim_{i\rightarrow\infty}x'_i\mbox{ for a Cauchy sequence }(x'_i)_i\mbox{ with }C(x'_i) \neq C(x) \mbox{ for all }i \}$. (This definition may in some cases allow the oracle to return a contrastive example with the same label as that of $x$. This is helpful when no point  $x'\in \mathcal{X}$ whose label differs from $C(x)$ attains the infimum of the values $d(x,x')$ for such $x'$, but mostly for complete metric spaces, where the limit of a Cauchy sequence is guaranteed to exist.\footnote{The definition of Cauchy sequence relies on a notion of metric. In our examples in continuous domains, we will use the $\ell_1$-metric both as this metric and as the function $d$ in the superscript of $\CS^d_{\min}$.})
    %We refer to this choice of contrast set as the \emph{minimum distance}\/ model.
    %\item $\CS^d_(x,C^*)=\arg\max_{x'\in\mathcal{X},C^*(x)=C^*(x')}d(x,x')$, i.e., the contrast set contains all instances farthest away from $x$, among those that have the same label under $C^*$ as $x$.
%    \item Fix $r > 0$. The contrast-oracle model given by
%$\CS^d_{\min, r}(x,C) := 
  %  \arg\min\{d(x,x')\ |\ x' \in \cX,\ C(x') \neq C(x),\ d(x, x') \leq r\}$
%\begin{equation} 
   % \CS^d_{\min, r}(x,C) := \begin{cases}
   % \arg\min\{d(x,x')\ |\ x' \in \cX: C(x') \neq C(x)\} & \exists x' \in \cX: C(x') \neq C(x), d(x, x') \leq r\\
 %   \emptyset & \text{o.w.}
 %   \end{cases}
%\end{equation}
%is referred to \emph{r-radius minimum distance model}.

\paragraph{Proximity model.} For our second contrast set mapping, the learner also selects a radius $r$ independently at each round, allowing it to ask for a contrastive example within a certain vicinity of the query point $x$. For discrete $\mathcal{X}$, we then define $\CS^d_{\prox}(x, r, C) = \{x' \in \mathcal{X} \mid C(x') \neq C(x), d(x, x')\leq r \}$, i.e., the contrast set contains all $x'$ $r$-close to $x$ that have the opposite label to $x$. For continuous $\cX$, a similar adaptation as for the minimum distance model yields $\CS^d_{\prox}(x, r, C) = \{x' \in \mathcal{X} |  d(x, x')\leq r,\ x'=\lim_{i\rightarrow\infty}x'_i$ for a Cauchy sequence $(x'_i)_i\mbox{ with }C(x'_i) \neq C(x) \mbox{ for all }i \}$. We refer to this choice of contrast set as the \emph{proximity}\/ model. Note that the oracle here need not choose a contrastive example at minimum distance; \emph{any}\/ example in  $\CS^d_{\prox}(x, r, C)$ is allowed.

Since these mappings do not depend on the version space, we dropped the argument $\cC_i$ in $\CS(x,C)$.

The following useful observations are easy to prove;  the proof of Remark~\ref{rem:prox} is given in the appendix.

\begin{remark}\label{rem:completeness}
    Let $\cC$ be a concept class over a complete space $\cX$ with metric $d$, $x \in \cX$, and $C \in \cC$. Suppose there is some $x' \in \cX$ with $C(x) \neq C(x')$ and $d(x, x') < \infty$. Then (i) $\CS^d_{\min}(x, C) \neq \emptyset$, and (ii) if $r \geq d(x, x')$ then $\CS^d_{\prox}(x, r, C) \neq \emptyset$.
\end{remark}
%\emph{Proof.} Obvious, since $(x')_i$ is a Cauchy sequence with $\lim_{i \rightarrow \infty} x' = x'$.
    %Note that $(x')_i$ is a Cauchy sequence with $\lim_{i \rightarrow \infty} x' = x'$. Thus $x' \in \CS^d_{\prox}(x, r, C)$ and this immediately implies (ii). Moreover, this implies $\{\mbox{Cauchy sequence }(x'_i)_i \mid C(x'_i) \neq C(x) \mbox{ for all }i\} \neq \emptyset$. Therefore, since every Cauchy sequence in $\cX$ has a limit, it follows $\CS^d_{\min}(x, C) \neq \emptyset$.
%\hfill$\Box$

%\FM{mention if CS' = 0 then CS = 0}
\begin{remark} \label{rem:cs}
Suppose that $\CS$ and $\CS'$ are mappings that assign to every
pair $(x,C) \in \cX \times \cC$ a subset of $\cX$. Suppose, for every $(x,C) \in \cX \times \cC$, we have (i) $\CS(x,C) \seq \CS'(x,C)$, and (ii)~$\CS(x,C)=\emptyset$ implies $\CS'(x,C)=\emptyset$.  Then $\cS_{\CS}(\cC) \le \cS_{\CS'}(\cC)$. Moreover, for any $\varepsilon$, $\cS_{\CS}(\cC,\varepsilon) \le \cS_{\CS'}(\cC,\varepsilon)$.
\end{remark}

%\emph{Proof.}
%The contrast oracle, viewed as an adversary of the learner, has no more freedom under $\CS$ than it has under $\CS'$.
%\hfill$\Box$

\begin{restatable}{remark}{remprox}
 \label{rem:prox}
     Let $\cC$ be a concept class over a complete space $\cX$ with metric $d$. Then $\cS_{\CS^{d}_{\min}}(\cC) \le \cS_{\CS^{d}_{\prox}}(\cC) \le \mathcal{S}[\MQ](\cC)$. 
     % The first inequality is Remark~\ref{rem:cs}, and 
\end{restatable}

\iffalse
\emph{Proof.} 
   The second inequality holds by definition. For the first inequality, consider any $r > 0$ and $(x,C) \in \cX \times \cC$. Due to Remark~\ref{rem:completeness}, $\CS^d_{\min}(x, C) = \emptyset$ iff there is no $x'$ with $C(x')\ne C(x)$. Moreover, $\CS^d_{\prox}(x, r, C) = \emptyset$ iff there are no points with different label converging to a point at distance at most $r$ from $x$. Therefore, either  $\CS^d_{\prox}(x, r, C) = \emptyset$, or $\CS^d_{\prox}(x, r, C) \supseteq \CS^d_{\min}(x, C) \neq \emptyset$. In the former case, the contrastive oracle in the minimum distance model also conveys that there are no points with different label converging to a point at distance at most $r$ from $x$. Consequently, the contrastive oracle in the proximity model does not provide extra information. In the later case, any contrastive example given by the contrastive oracle in the minimum distance model can be also given by the contrastive oracle in the proximity model. Again, the contrastive oracle in the proximity model does not provide extra information.
   % he same reasoning as for Remark~\ref{rem:cs} applies.
\hfill$\Box$
\fi

%~We conclude this section with a relationship between  $\cS_{\CS^{d}_{\prox}}$ and $\cS_{\CS^{d}_{\min}}$ for finite $\cX$.

When $\cX$ is finite, we can extend Remark~\ref{rem:prox} as follows.

\begin{proposition}\label{prop:min-prox}
    Fix $\mathcal{X}$ and $d:\mathcal{X}\times\mathcal{X}\rightarrow\mathbb{R}^{\ge 0}$. Let $S_d(x) = \{d(x,x')\mid x' \in \cX \setminus {x}\}$ and $s_d = \max_{x \in \cX} |S_d(x)|$. (Note that $s_d\le|\cX|-1$ for finite $\cX$). 
Then $\cS_{\CS_{\prox}^d}(\cC) \leq \lceil \log(s_d) \rceil \cdot \cS_{\CS_{\min}^d}(\cC)$.
    % Then $\cS_{\CS^{d}_{\prox}}(\cC) \le \log(|\mathcal{X}|)\cdot\cS_{\CS^{d}_{\min}}(\cC)$. 
\end{proposition}

\emph{Proof.}
Let $L$ learn any $C \in \cC$ with  $\le \cS_{\CS^{d}_{\min}}(\cC)$ queries to the minimum distance oracle w.r.t.\ $d$. 
We construct a learner $L'$ with access to the proximity oracle, using $L$ as a subroutine. Suppose $L$ poses a query $x$. Let $r_{\min} = \min_{x' \in \CS^d_{\min} (x, C)} d(x, x')$. First, $L'$ sorts all numbers in $S_d(x)$ in increasing order and determines the median $r$ of that list. Then $L'$ poses the query $(x, r)$. The contrastive example returned will be  $\omega$ iff $d(x, x') < r_{\min}$. Hence, with a binary search using $\lceil \log(s_d) \rceil$ queries, $L'$ can determine an $x'$ with opposite label to $x$ and $d(x, x') = r_{\min}$. This $x'$ is among the admissible answers to the query by $L$. Thus, $L'$ makes $\le \lceil \log(s_d) \rceil$ queries for every query that $L$ makes.
% For any $C \in \cC$, and let $[(x_1,y_1),(x'_1,y'_1),\mathcal{C}_2],\ldots,[(x_n,y_n),(x'_n,y'_n),\mathcal{C}_{n+1}]$ be minimal interaction sequence for identifying $C$, where $n \leq \cS_{\CS^{d}_{\min}}(\cC)$. 
\hfill$\Box$

%In essence, the proof idea for Proposition~\ref{prop:min-prox} involves a  learner actively choosing the radius $r$ via binary search until the target decision boundary falls within an $r$-ball around a known instance $x$. 
%The binary search approach in this proof relies on the learner's knowledge of an upper bound on $d(x,x')$ for $x,x'\in\mathcal{X}$. Hence, it does not carry over to cases in which no such upper bound exists. 

\section{Sample Complexity Under Various Metrics}

This section illustrates our notions of learning from contrast oracles under various metrics $d$. %as choices for $d$, such as the $\ell_1$-norm for concepts over real numbers, the Hamming distance for Boolean functions, and the discrete metric (to be defined in Section~\ref{ssec:discrete}) as a worst-case example of a metric.
We begin with two examples on the \textbf{$\ell_1$-metric}.

%\subsection{The $\ell_1$-distance}

%Throughout this subsection, $d$ is the $\ell_1$-metric. We provide two examples.
%As a first example, consider 1-sided threshold functions in 1 dimension.
\begin{table}[t] 
\centering
\begin{tabular}{|l|c|c|c|}
        \hline
& without $\CS$&$\CS^d_{\prox}$&$\CS^d_{\min}$ (even for exact identification)\\\hline
(a) thresholds& $\Theta(\log \frac{1}{\varepsilon})$&  $\Theta(\log \frac{1}{\varepsilon})$ &1\\\hline
(b)  rectangles&$\Theta(k\log(\frac{1}{\varepsilon}))$&$\Omega(\log(\frac{1}{\varepsilon}))$,  $O(\log(\frac{k}{\varepsilon}))$&2\\\hline
\end{tabular}
\iffalse
        \begin{tabular}{|l|l|l|}
        \hline
                                                                                     & (a) thresholds 
                                                                                     & (b)  rectangles\\ \hline
        without $\CS$                                               & $\Theta(\log \frac{1}{\varepsilon})$  &$\Theta(k\log(\frac{1}{\varepsilon}))$  \\ \hline
        $\CS^d_{\prox}$                                         &  $\Theta(\log \frac{1}{\varepsilon})$ & $\Omega(\log(\frac{1}{\varepsilon}))$,  $O(\log(\frac{k}{\varepsilon}))$ \\ \hline
        $\CS^d_{\min}$          & $1$  & $2$                          \\ \hline
        %r-radius minimum distance model & $\frac{1}{r} + 1$ & $\log \frac{1}{r} + 1$ \\ \hline
        % minimum distance model (learner doesn't know the model) & $O(1)$                             & $O(\log \frac{1}{\varepsilon})$ \\ \hline
        \end{tabular}
        \fi
        \caption{Asymptotic sample complexity when $\varepsilon$-approximating (a) one-sided threshold functions, and (b) axis-aligned rectangles in $k$ dimensions, for various learning models, with the $\ell_1$-metric.} %The results for $\CS^d_{\min}$ even hold for exact identification.}
        \label{tab:thresholds-ex}
        \end{table}
\begin{example} \label{exmp:thresholds}
        Consider the class of one-sided threshold functions $\mathbbm{1}\{x \leq \theta^*\}$ for $\theta^* \in \cX = [0, 1]$. %, together with the $\ell_1$-distance as metric $d$. %, with the uniform distribution as marginal distribution. 
        Table~\ref{tab:thresholds-ex}(a) displays the asymptotic sample complexity of various models of $\varepsilon$-approximate learning. %The rows represent different models of contrastive learning. Moreover, two columns represent (i) learner actively choose samples (ii) samples are drawn randomly from marginal distribution. 
        The result on  learning without contrastive examples is known %, see, e.g., 
        \cite{Settles12}.
        By a straightforward adversary argument, 
    the sample complexity of the proximity model is in 
    $\Omega(\log 1/\varepsilon)$. As learning without contrast oracle is no stronger than learning in the proximity model, both models here have the same asymptotic complexity.
    In the minimum-distance model, irrespective of the first queried instance, the contrastive example will be  $\theta^\star$. Thus, the target can be identified (even exactly!) with one query.
\end{example}
\begin{example}\label{exmp:rectangles} Consider the class of all $k$-dimensional axis-aligned rectangles over $\cX = [0, 1]^k$, with the $\ell_1$-distance as metric. Table~\ref{tab:thresholds-ex}(b) displays the asymptotic sample complexity of various models of $\varepsilon$-approximate learning.
%of active learning (i) without contrastive examples (membership query) (ii) in the proximity model with $\ell_1$-distance, and (iii) in the minimum distance model with with $\ell_1$-distance.
%\end{example}
%\emph{Proof.}
The result on learning without contrastive examples is known, see, e.g., \cite{Hanneke09}. Let $C^\star$ be a target rectangle with the corner of lowest $\ell_1$-distance from $\vec 0=(0,\ldots,0)$ being $\vec x$, and the diagonally opposed corner being $\vec y$.
\\
In the minimum-distance model, let a learner first query $\vec 0$. If $C^*(\vec 0)=1$ then clearly $\vec x = \vec 0$. If $C^*(\vec 0)=0$ then the contrastive example $x'$ will be $\vec x$.  Similarly, by querying $\vec 1$, the learner can determine $\vec y$. Thus, $\cS_{\CS^d_{\min}}\le 2$. Also, for determining $C^*$, at least two distinct positively labelled instances are required, which is impossible to acquire with one query. %This completes the proof.
\\
To verify the upper bound in the proximity model, note that $d(\vec x, \vec 0) \leq k$ for any $\vec x$. The learner starts by querying $x_1 = \vec 0$ and $r_1 = \frac{k}{2}$. The contrastive example will be $\omega$ iff $r_1 < d(\vec x, \vec 0)$. Thus, the learner determines whether $d(\vec x, \vec 0) \leq r_1$ or not. Learning continues with a binary search until finding a radius $r_t$ such that $d(\vec x, \vec 0) \leq r_t \leq d(\vec x, \vec 0) + \varepsilon / 2$ for $ t \geq \log(2 k / \varepsilon) + 1$. Let $x'_t$ be the contrastive example. We derive
$d(x'_t, \vec x) \leq d(x'_t, \vec 0) - d(\vec x, 0) \leq r_t - d(\vec x, \vec 0) \leq \frac{\varepsilon}{2}$ (noting that $d(x'_t, \vec 0) - d(\vec x, 0)>0 $ at step $t$).
Similarly with $\log(2k / \varepsilon) + 1$ samples the learner can find a $\vec z \in [0, 1]^k$ such that $d(\vec z, \vec y) \leq \varepsilon/2$. Therefore, the rectangle with bottom left corner $x'_t$ and top right corner $\vec z$ has error less than $\varepsilon$. %This completes the proof.
\\
%It remains to prove the lower bound for the proximity model. For $i \in [k]$, let $\vec{1}_i$ be the vector with $i$th component 1 and all other components zero. Note that for $k$-dimensional rectangles, the information provided by any query $(x, r)$ is equivalent to the information from plain membership queries $x \pm r \vec 1_1$, $x \pm r \vec 1_2$, $\cdots$, $x \pm r \vec 1_k$. Hence the sample complexity of the proximity model is asymptotically lower-bounded by the sample complexity of the membership query model divided by $k$. This would be $k \log (1/\varepsilon) /k = \log 1/\varepsilon$.
The lower bound on $\mathcal{S}_{\CS^d_{\prox}}(\cC,\varepsilon)$ is inherited from that in Example~\ref{exmp:thresholds}, using the same argument.
%\hfill$\Box$
\end{example}

%\subsection{The Hamming Distance}

%In this section, we consider 
The next few examples concern concept classes over the Boolean
domain $B_m = \{0,1\}^m$ and learning from a contrast oracle in the
minimum distance model with respect to the \textbf{Hamming distance as metric}. 
For $I \seq [m]$, we denote by $\vec{1}_I$ the Boolean vector with $1$s
in positions indexed by $I$ and $0$s elsewhere; let $\vec{1}:=\vec{1}_{[m]}$. The notation $\vec{0}_I$ and $\vec{0}$ is
understood analogously. 

The next two toy examples will show that a certain metric (here the Hamming distance) may be suitable in the context of a specific concept class, but will become misleading in the minimum distance model when a slight representational change is made to the concept class. These toy examples are merely meant to illustrate the power and limitations of our abstract model in extreme situations.

\begin{restatable}{example}{exmmonfirst}\label{exmp:monomials1}
Let $\mathcal{X}=\{0,1\}^m$. Let $\mathcal{C}^m_{\pmon}$ consist of all monotone monomials, i.e., logical formulas of the form $v_{i_1}\wedge \ldots\wedge v_{i_k}$ for some pairwise distinct $i_1,\ldots,i_k$ and some $k\in\{0,\ldots,m\}$. The concept associated with such a formula contains the boolean vector $(b_1,\ldots,b_m)$ iff $b_{i_1}=\ldots =b_{i_k}=1$ (where the empty monomial is the constant 1 function). 
Let $d$ be the Hamming distance. 
% and let \[\CS(x,C^*)=\arg\min_{x'\in\mathcal{X},C^*(x)\ne C^*(x')}d(x,x')\,.\]
Then:  %the following statements hold.
\iffalse
$$
(1)\ \mathcal{S}[\MQ](\mathcal{C}^m_{\pmon}) = m; \ \ (2)\ \mathcal{S}_{\CS^d_{\min}}(\mathcal{C}^m_{\pmon})=1;\ \ (3)\ \mathcal{S}_{\CS^d_{\prox}}(\mathcal{C}^m_{\pmon})=\Theta(\log m)\,.
$$
\fi
\smallskip
\centerline{
(1)  $\mathcal{S}[\MQ](\mathcal{C}^m_{\pmon}) = m$.\ \ \ 
    (2) $\mathcal{S}_{\CS^d_{\min}}(\mathcal{C}^m_{\pmon})=1$.\ \ \   (3) $\mathcal{S}_{\CS^d_{\prox}}(\mathcal{C}^m_{\pmon})=\Theta(\log m)$.}
\end{restatable}
 
 \emph{Proof.} \emph{(Sketch.)} %Suppose a membership query learner queries each boolean vector that has a 1 in every component except a single component $v_j$. The answer to such query will reveal whether or not $v_j$ is in the target formula. Clearly, these $m$ queries determine the target concept. The standard information-theoretic lower bound makes this learner optimal in terms of the worst-case number of membership queries on $\mathcal{C}^m_{\pmon}$. 
% \FM{\cite{abasi2014exact} in Lemma 2 has proved that it is at-least $\log |\cC| - 1 = m - 1$ for any randomized learner that want to identify $C^\star$ with probability at least 3/4. But it also mentions that it is commonly assumed that any deterministic algorithm requires atleast $\log |\cC|$ samples. But they didn't find any proof and they believe it's ``folklore'' result, and I think this is also correct, but can't find reference} 
(1) is folklore. For (2), a learner will first query $\vec 0$, which has the label 0 unless the target is the empty monomial. The unique positively labeled example of smallest Hamming distance to the all-zeroes vector has ones exactly in $v_{i_1}$,\ldots,$ v_{i_k}$, where the target concept is $v_{i_1}\wedge \ldots\wedge v_{i_k}$. Thus, a single query identifies the target. The upper bound in (3) follows from Proposition~\ref{prop:min-prox}, while the lower bound is obtained via case distinction.  Details are in the appendix.
 \hfill$\Box$

 Now, we will see how a small syntactic change from  $\mathcal{C}^m_{\pmon}$ to a class $\mathcal{C}'_{\pmon}$ (adding an $(m+1)$st component to the boolean vectors) can make contrastive examples useless (see appendix for details). 
 
 \begin{restatable}{example}{exmmon}\label{exmp:monomials2}
 Let $\mathcal{X}'=\{0,1\}^{m+1}$. The concept class $\mathcal{C}'_{\pmon}$ over $\mathcal{X}'$ is defined to contain, for each $C\in\mathcal{C}^m_{\pmon}$, the concept $C'\in\mathcal{C}'_{\pmon}$ constructed via $C'(b_1,\ldots,b_m,0)=C(b_1,\ldots,b_m)$ and $C'(b_1,\ldots,b_m,1)=1-C(b_1,\ldots,b_m)$.
 No other concepts are contained in $\mathcal{C}'_{\pmon}$. 
 Let $d$ be the Hamming distance.
 % and let \[\CS(x,C^*)=\arg\min_{x'\in\mathcal{X}',C^*(x)\ne C^*(x')}d(x,x')\,.\]
 Then $\mathcal{S}[\MQ](\mathcal{C}'_{\pmon})=\mathcal{S}_{\CS^d_{\min}}(\mathcal{C}'_{\pmon})=\mathcal{S}_{\CS^d_{\prox}}(\mathcal{C}'_{\pmon})=m$.
 %Then the following statements hold.
%\begin{enumerate}
 %   \item $\MQ(\mathcal{C}'_{\pmon})=m$.
 %   \item $\mathcal{S}_{\CS^d_{\min}}(\mathcal{C}'_{\pmon})=m$.
%    \item  $\mathcal{S}_{\CS^d_{\prox}}(\mathcal{C}'_{\pmon})=m$.
%\end{enumerate}
\end{restatable}

\iffalse
 \emph{Proof.}
 A membership query learner using $m$ queries will work as described in the proof of Example~\ref{exmp:monomials1}, yet will leave the $(m+1)$st component of every queried vector equal to 0. Thus, the learner will identify which vectors of the form $(b_1,\ldots,b_m,0)$ belong to the target concept. The latter immediately implies which vectors of the form $(b_1,\ldots,b_m,1)$ belong to the target concept. Again, by a standard information-theoretic argument, fewer membership queries will not suffice in the worst case. Thus $\mathcal{S}[\MQ](\mathcal{C}'_{\pmon})=m$.
 
 Upon asking any query $(b_1,\ldots,b_m,b_{m+1})$, a learner in the minimum distance model will receive the correct answer, plus the contrastive example $(b_1,\ldots,b_m,1-b_{m+1})$. This contrastive example does not provide any additional information to the learner. Hence, the learner has to ask as many queries as an $\MQ$-learner, i.e., $\mathcal{S}_{\CS^d_{\min}}(\mathcal{C}'_{\pmon})=m$.
 
  Finally, according to Remark~\ref{rem:prox}, $\mathcal{S}_{\CS^d_{\min}} \leq \mathcal{S}_{\CS^d_{\prox}} \leq \mathcal{S}[\MQ]$. Thus, $\mathcal{S}_{\CS^d_{\prox}}(\mathcal{C}'_{\pmon})=m$.
\hfill$\Box$
\fi

While the Hamming distance is useless for the class $\mathcal{C}'_{\pmon}$ in Example~\ref{exmp:monomials2}, other functions $d$ might still allow for efficient learning in this setting. Such functions, however, might be rather ``unnatural'' as a notion of distance (arguably though, the class $\mathcal{C}'_{\pmon}$ in Example~\ref{exmp:monomials2} is also somewhat ``unnatural''). We will hence focus most of our study on intuitively ``natural'' functions $d$. %, without being able to define what exactly a natural metric is. 
When learning Boolean functions,  %(represented as concept classes over vectors of true/false assignments to their variables), 
we consider Hamming distance a natural choice.

Let $\MONOMIALS^m$ denote the class 
of monomials over the Boolean variables $v_1,\ldots,v_m$ (including the
empty monomial which represents the constant-1 function). We assume
that each Boolean variable occurs at most once (negated or not negated)
in a monomial. Since monomials like $v_i \wedge\bar v_i$  are thus excluded, the constant-0 function does not belong to $\MONOMIALS^m$. A concept class dual to $\MONOMIALS^m$ is the class of clauses over the Boolean
variables $\{v_1,\ldots,v_m\}$ (including the empty clause which represents
the constant-0 function but excluding clauses like $v_i \vee \bar v_i$
which represent the constant-1 function). We denote this class by $\CLAUSES^m$, and obtain the following result, the full proof of which is given in the appendix.

\noindent
%Lemma~\ref{lem:monomials} and Corollary~\ref{cor:clauses} are subsumed by the following (slightly) stronger result:

\begin{restatable}{theorem}{coralternation}\label{cor:alternation1}
$\mathcal{S}_{\CS^d_{\min}}(\MONOMIALS^m \cup \CLAUSES^m)=2$, where $d$ is the Hamming distance.
%The class $\MONOMIALS^m \cup \CLAUSES^m$ can be learned from a contrast oracle in the minimum-distance model at the expense of $2$ queries.
\end{restatable}

\emph{Proof.} \emph{(Sketch.)}
One first constructs a learner $L_\wedge$ for $\MONOMIALS^m$, as in the proof of Example~\ref{exmp:monomials1}, and then a learner $L_\vee$ for $\CLAUSES^m$ by dualization from $L_\wedge$. It suffices to describe a learner $L$ that maintains
simulations of $L_\wedge$ and $L_\vee$ and \emph{either} comes in both
simulations to the same conclusion about the target
concept
\emph{or} realizes an inconsistency in one of the simulations,
which can then be aborted. %Details are given in the appendix.
\hfill$\Box$

For $\mathcal{S}_{\CS^d_{\prox}}$, an upper bound is obtained from  Proposition~\ref{prop:min-prox} and Theorem~\ref{cor:alternation1}, noting that the Hamming distance can take only $m$ distinct values for pairs $(x,x')$ with $x\ne x'$.

\begin{restatable}{corollary}{proxmonclaus}
$\mathcal{S}_{\CS^d_{\prox}}(\MONOMIALS^m)\cup\mathcal{S}_{\CS^d_{\prox}}(\CLAUSES^m)\le 2 \lceil\log m\rceil$, where $d$ is the Hamming distance.
    %$\MONOMIALS^m$, $\CLAUSES^m$ can be learned from a contrast oracle in the proximity model at the expense of $2 \log m$ queries.
\end{restatable}
\iffalse
\emph{Proof.}
    We prove the claim for $\MONOMIALS^m$. A learner $L$ first queries $(\mathbf 1, \lfloor\frac{m}{2}\rfloor)$. With a binary search as in the proof of Example~\ref{exmp:monomials1}.3, $L$ learns $|I|$ with $\log m$ queries. Now a query $(\mathbf 1, |I|)$ is equivalent to querying $\mathbf 1$ in the minimum distance model. Similarly, $L$ queries $(\mathbf 0, \lfloor\frac{m}{2}\rfloor)$, and finds $|J|$ with $\log m$ queries; then the query $(\mathbf 0, |J|)$ is equivalent to querying $\mathbf 0$ in the minimum distance model. Using Proposition~\ref{lem:monomials} this concludes the proof.
\hfill$\Box$
\fi

\iffalse
and the $\ell_1$-metric
\[ d_1(x,y) = \sum_{i=1}^{k}|y_i-x_i| \enspace , \]
provided that $\cX \seq \reals^k$.

We have already seen some illustration of contrastive learning using the $\ell_1$-metric in Examples~\ref{exmp:thresholds} and \ref{exmp:rectangles}. The next subsection discusses the discrete metric in more detail.
\fi

We conclude this subsection with an example of a somewhat natural concept class for which the minimum distance model under the Hamming distance is not very powerful.

\begin{example}\label{exmp:parity}
Let $\mathcal{X}=\{0,1\}^m$ and let $\cC_{\parity}$ be the class of parity functions, i.e., logical formulas of the form $v_{i_1} \oplus  \ldots \oplus  v_{i_k}$ for some pairwise distinct $i_1,\ldots,i_k$ and some $k\in\{0,\ldots,m\}$. 
% The concept associated with such a formula contains the boolean vector $(b_1,\ldots,b_m)$ iff $b_{i_1}=\ldots =b_{i_k}=1$ (where the empty monomial is the constant 1 function). 
Then $\mathcal{S}[\MQ](\cC_{\parity})=m$ and $m - 1 \leq \mathcal{S}_{\CS^d_{\min}}(\cC_{\parity})=\mathcal{S}_{\CS^d_{\prox}}(\cC_{\parity}) \leq m$.
\end{example}
\emph{Proof.}
    $\mathcal{S}[\MQ](\cC_{\parity})=m$ is trivial, and immediately yields  $ \mathcal{S}_{\CS^d_{\min}}(\cC_{\parity})\leq m$, $\mathcal{S}_{\CS^d_{\prox}}(\cC_{\parity}) \leq m$. Suppose the target concept is $c^* = v_{i_1} \oplus  \ldots \oplus  v_{i_k}$.  For any instance $\vec x$, we have $d(\vec x, \vec x') = 1$ and $c^*(\vec x') \neq c^*(\vec x')$, where $\vec x'$ results from $\vec x$ by flipping the value of component $i_1$. %Obviously $d(\vec x, \vec x') = 1$ and $c^*(\vec x') \neq c^*(\vec x')$. 
    When always using a query radius $r_t = 1$, the proximity model is thus equivalent to the minimum distance model. The lower bound is witnessed by an oracle that contrasts any $\vec x_t$ with the example $\vec x'_t$ resulting from $\vec x_t$ by flipping the component $i_1$. This provides no information about variables other than $v_{i_1}$. % This indicates that minimum distance model and  are equivalent.
    % Then, note that if there exists a contrastive example, the vector with $i_1$ index flipped will be a contrastive example. Therefore, there will always exist one with distance 1. Thus, proximity model is equivelant with minimum distance model. Then imagine the contrastive example is always the one with $i_1$ flipped. Then contrastive example only give knowledge of $v_{i_1}$. Thus, or not the sample complexity will be $m - 1$. 
\hfill$\Box$

%\subsection{The Discrete Metric}\label{ssec:discrete}
%\FM{show that condition of Remark 4 is satisfied - this should hold even for complete spaces. Our argument below ignores complete spaces but can be modified.}
We conclude this section with the \textbf{discrete metric $d_0$}, defined by $d_0(x,x')=0$ if $x= x'$, and $d_0(x,x')=1$ if $x\ne x'$.
%\[
%d_0(x,y) = \left\{ \begin{array}{ll}
 %           0 & \mbox{if $x=y$} \\
  %%      \end{array}\right.
%\]
%Under this metric, %the value $1$ equals the
%maximum and also the minimum distance between two different
%instances in $\cX$. For this reason, 
%the following holds for
Now $\CS^{d}_{\min}(x,C)\subseteq \CS^{d_0}_{\min}(x,C)$ for every $(x,C) \in \cX \times \cC$ and every $d:\mathcal{X}\times\mathcal{X}\rightarrow\mathbb{R}^{\ge 0}$. Moreover,  $\CS^{d}_{\min}(x,C)=\emptyset$ implies  $\CS^{d_0}_{\min}(x,C)=\emptyset$.
%$\CS^{d}_{\min}(x,C)=\{x' \in \cX\mid C(x') \neq C(x),\ 
%d(x,x') = \min_{x'': C(x'') \neq C(x)}d(x,x'')\}$. The latter is a subset of $\{x' \in \cX\mid C(x') \neq C(x)\}=\CS^{d_0}_{\min}(x,C)$.
\iffalse
\begin{eqnarray*}
\CS^{d_0}_{\min}(x,C) & = & \{x' \in \cX\mid C(x') \neq C(x)\} \\
& \supseteq & \left\{x' \in \cX\mid C(x') \neq C(x),\ 
d(x,x') = \min_{y: C(y) \neq C(x)}d(x,y)\right\} = \CS^{d}_{\min}(x,C) \,. 
\end{eqnarray*}
\fi
%In combination with 
Remark~\ref{rem:cs} thus yields:

\begin{corollary} \label{cor:maximizer-d0}
For every concept class $\cC$, the term $\cS_{\CS^{d}_{\min}}(\cC)$
is maximized,
if the underlying function $d:\mathcal{X}\times\mathcal{X}\rightarrow\mathbb{R}^{\ge 0}$ equals the discrete
metric $d_0$.
\end{corollary}
%This comes, of course, at no surprise because the contrast oracle
%(viewed as an adversary of the learner) will get the most degrees
%of freedom if we set $d = d_0$. 

In what follows, if $L$ is a list of oracles, we use $\mathcal{S}[L]$ to denote the sample complexity of learning with access to all oracles in $L$. Let $\EX^+$ be an oracle returning
a positive example $x$, i.e., an $x \in \cX$ with $C^*(x)=1$ (or a dummy response if no such $x$ exists). When called several times, the oracle may repeatedly
present the same example. Let $\EX^-$ be the
analogous oracle for a negative example.
The following result says that, under the metric $d_0$, the contrast oracle does not
provide significantly more information than the membership oracle 
(although, for some special concept classes, this small extra information can make a big difference). The proof is given in the appendix. %demonstrates how to simulate calls to the contrast oracle with calls to the other three oracles, and vice versa.

\begin{restatable}{theorem}{thmplusminusmq} \label{rem1:d0}
%In the minimum-distance model with metric $d_0$, the following holds:
$\mathcal{S}[\EX^+,\EX^-,\MQ](\cC)-2\ \le\ \cS_{\CS^{d_0}_{\min}}(\cC)\ \le\ \mathcal{S}[\EX^+,\EX^-,\MQ](\cC)$.
\end{restatable}

\begin{example}
Consider $\cC $ to be the class of all singleton concepts over a finite domain $\cX$ where $n:=|\cX|\ge2$. It is known
that $\mathcal{S}[\MQ](\cC)=n-1$.  But a single call of the oracle $\EX^+$
reveals the identity of the target concept. It follows that
$\mathcal{S}[\EX^+](\cC)\ =\ \mathcal{S}[\EX^+,\EX^-,\MQ](\cC)\ =\ \cS_{\CS^{d_0}_{\min}}(\cC)\
=\ 1$.
\end{example}

\iffalse
\subsection{The $\ell_1$-Metric}

\begin{example}
Consider axis-aligned boxes on a grid, i.e., $\cC = \BOXES^{k,n}$ for $n\ge4$ and $k\ge1$, which is the class
of concepts of the form $[i_1:j_1] \times\ldots\times [i_k:j_k]$ where
$1 \le i_\ell \le j_\ell \le n$ for $\ell=1,\ldots,k$. We claim that
$\cS_{CS_{min}^{d_1}}(\cC) = 2$.

It is easy to see that a single query is not enough. To see that $2$ queries are enough,
let $C^* = [i_1:j_1] \times\ldots\times [i_k:j_k]$ be the target box.
%The central observations are as follows:
%\begin{itemize}
%\item
Once two opposite corners of $C^*$ are known (e.g., the
corner $x^* = (i_1,\ldots,i_k)$ with the smallest coordinates
and the corner and $y^* = (j_1,\ldots,j_k)$ with the largest
coordinates), the whole target box $C^*$ is known.

%\item
Via the query point $x = (1,\ldots,1)$, the learner gets to know
the corner $x^*$: if $x^* = x$, then the contrast oracle must return
the label $1$ for $x$; otherwise, if $x^* \neq x$, then the contrast
oracle returns the label $0$ for $x$ and also the point $x^*$ because
the latter is the point from $C^*$ with the smallest $d_1$-distance
to the query point $x$.
%\item

For reasons of symmetry, the following is true as well: via the query
point $y = (n,\ldots,n)$, the learner gets to know the
corner $y^* = (j_1,\ldots,j_k)$.
%\end{itemize}
Thus two queries suffice, indeed. 
\end{example}

\fi

\section{Lower Bounds}
\label{sec:sd}

In this section, we will derive some lower bounds on the sample complexity in our models of learning from contrastive examples. All our results here are for finite instance spaces.

%\textcolor{red}{Here I first plan to introduce Farnam's lower bound only for single metrics, since at this point we have not yet talked about sequences of metrics. Later on we can say that the result obviously goes through even for sequences of metrics.}

\subsection{A Metric-Independent Lower Bound}

Self-directed learning \cite{GoldmanS94} is a model of online learning, in which the sequence of data to be labeled is chosen by the learner as opposed to the environment. The learner gets immediate feedback on every single prediction, which allows it to adapt its data selection to the version space sequentially. It is then assessed in terms of the number of incorrect predictions it makes across all data instances in $\cX$; as usual, we consider the worst case over all concepts in the underlying class~$\cC$. 

\begin{definition}[\cite{GoldmanS94}]
    Let $\cC$ be a concept class over a finite domain~$\cX$, $n=|\cX|$. A self-directed learner $L$ for $\cC$ interacts with an oracle in $n$ rounds. In each round, $L$ selects a not previously selected instance $x\in\cX$ and predicts a label $b$ for $x$. The oracle knows the target $C^*\in\cC$ and returns the label $C^*(x)$; if $b\ne C^*(x)$, the learner incurs a mistake. $L$ proceeds to the next round until all instances in $\cX$ have been labeled. The cost of $L$ on $C^*$ is the number of mistakes made in these $n$ rounds. The cost of $L$ on $\cC$ is the maximum cost of $L$ on any $C\in\cC$.
\\
    The self-directed learning complexity of $\cC$, denoted by $\SD(\cC)$, is the smallest number $c$ such that there exists a self-directed learner for $\cC$ whose cost on $\cC$ is $c$. 
\end{definition}

This complexity notion was analyzed relative to other learning-theoretic parameters \cite{Ben-DavidEK95,DoliwaFSZ14}, and extended to infinite concept classes and multi-class settings \cite{DevulapalliH24}. Interestingly, $\SD$ can be used to derive a metric-independent lower bound on the sample complexity of the minimum-distance model.

\iffalse
\begin{theorem}\label{thm:sd}
    Let $\mathcal{X}$ be a finite domain and $\cC$ a concept class over $\mathcal{X}$. For any choice of $d:\mathcal{X}\times\mathcal{X}\rightarrow\mathbb{R}^{\ge 0}$, we have %$\cS_{\CS^{d}_{\min}}(\cC) \geq 1$ in case $\operatorname{SD}(\cC)=1$, and
     $\cS_{\CS^{d}_{\min}}(\cC) \geq \lceil\operatorname{SD}(\cC)/2\rceil$.
\end{theorem}
\fi

\begin{theorem}\label{thm:sd}
    If $\cC$ is defined over a finite $\mathcal{X}$, and $d:\mathcal{X}\times\mathcal{X}\rightarrow\mathbb{R}^{\ge 0}$, then %$\cS_{\CS^{d}_{\min}}(\cC) \geq 1$ in case $\operatorname{SD}(\cC)=1$, and
     $\cS_{\CS^{d}_{\min}}(\cC) \geq \lceil\operatorname{SD}(\cC)/2\rceil$.
\end{theorem}
\emph{Proof.} Fix $d:\mathcal{X}\times\mathcal{X}\rightarrow\mathbb{R}^{\ge 0}$, and let $L$ be a learner with access to a minimum-distance oracle w.r.t.\ $d$, which makes at most $\cS_{\CS^{d}_{\min}}(\cC)$ queries to identify any $C\in\mathcal{C}$. 
We construct a self-directed learner $L'$ for $\mathcal{C}$ which calls $L$ as a subroutine. If $L$ poses a query for $x$, then $L'$ predicts 0 on $x$. %(the actual prediction does not matter for this proof). 
After receiving feedback, $L'$  passes on the correct label $b$ for $x$ to $L$. In addition, $L'$ must provide a contrastive example for $x$, which is obtained as follows: $L'$ sorts all instances in $\mathcal{X}$ in increasing order w.r.t.\ their distance from $x$ as measured by $d$. It then predicts the label $b$ for instances in this ordered list until it observes the first mistake, say for $x'$. Then $x'$ must be a closest instance to $x$ with a label opposite from $b$. Hence, $x'$ is passed on to $L$ as a contrastive example. If no such $x'$ is found, then $L'$ gives the response $\omega$ to $L$. This proceeds until all instances in $\mathcal{X}$ have been labeled. 

Each time $L'$ calls $L$, it incurs at most two mistakes, namely one for its prediction on the instance $x$ queried by $L$, and one for its prediction on the contrastive instance $x'$. In total, thus $L'$ makes at most $2t$ mistakes, where $t$ is the number of queries posed by $L$. We obtain the desired bound. \hfill$\Box$

%\FM{new remark}

\begin{remark}\label{rem:vcd}
      For each $m\in\mathbb{N}$, there is a concept class $\cC_m$ with $\VCD(\cC_m)=2$ and $\SD(\cC_m) \geq m$ \cite{10.5555/646945.712381}. In particular, there is no relationship between $\VCD$ and $\SD$. Thus, no matter which function $d: \cX \times \cX \rightarrow \reals^{\geq 0}$ is chosen, there exists no upper bound on $\cS_{\CS^d_{min}}$ as a function of $\VCD$.
\end{remark}

We will see multiple cases in which the lower bound provided by Theorem~\ref{thm:sd} is not met. As a small warm-up example, there are finite classes of VC dimension 1 for which $\cS_{\CS^{d}_{\min}}(\cC) =2$, while every class of VC dimension 1 satisfies $\SD=1$ (see the appendix for proof details):

\iffalse
\begin{table}[] 
\caption{A concept class $\cC$ with $\SD(\cC) = \VCD(\cC) = 1$, but $\cS_{\CS^d_{\min}}(\cC) \geq 2$ for any metric $d$.}
\centering
\begin{tabular}{|l|l|l|l|}
\hline
 & $x_1$ & $x_2$ & $x_3$ \\
\hline
$C_1$                & 1     & 0     & 1     \\
\hline
$C_2$                & 0     & 0     & 1     \\
\hline
$C_3$                & 0     & 1     & 1     \\
\hline
$C_4$                & 0     & 1     & 0    \\
\hline
\end{tabular}
\label{tab:vc-1-cs-2-ex}
\end{table}
\fi

\begin{restatable}{example}{exmpvcd}
    Let $\cX=\{x_1,x_2,x_3\}$. Let $\cC$ consist of the concepts $\{x_1,x_3\}$, $\{x_3\}$, $\{x_2,x_3\}$, and $\{x_2\}$. Then $\cC$ satisfies $\SD(\cC) =\VCD(\cC) = 1$, but $\cS_{\CS^d_{\min}}(\cC) \geq 2$ for any $d:\cX \times \cX \rightarrow \reals^{\geq 0}$.
\end{restatable}
\iffalse
\emph{Proof.}

By case distinction---see the appendix for details.

Regardless of how the distance function $d$ is chosen, the learner will get to 
see no more than two labeled examples, one labeled $0$, the other one labeled $1$.
It suffices therefore to show that, for each choice of two instances, the two
opposing labels for them can be chosen such that the resulting version space 
contains at least two distinct concepts. This can be achieved as follows:
\begin{itemize}
\item
Given $x_1,x_2$, choose label $0$ for $x_1$, and $1$ for $x_2$.
The resulting version space is $\{C_3,C_4\}$.
\item
Given $x_1,x_3$, choose label $0$ for $x_1$, and  $1$ for $x_3$.
The resulting version space is $\{C_2,C_3\}$.
\item
Given $x_2,x_3$, choose label $0$ for $x_2$, and  $1$ for $x_3$.
The resulting version space is $\{C_1,C_2\}$.
\end{itemize} 
In any case, the resulting version space 
is of size $2$.
\hfill$\Box$
\fi

However, for classes of VC dimension 1, we cannot increase the gap between $\SD$ and $\min_d\cS_{\CS^d_{\min}}$: % any further, as the next theorem shows.

\iffalse
The following useful fact was noted by \cite{DoliwaFSZ14}.

\begin{lemma}\label{lem:vcd-1-column}
    Let $\cC$ be any concept class with $\VCD(\cC) = 1$. Then there is at least one $(x,b)\in\mathcal{X}\times\{0,1\}$ such that there exists exactly one $C\in\mathcal{C}$ with $C(x)=b$.
\end{lemma}
\fi

\begin{restatable}{theorem}{vcdOne}\label{thm:vcd1}
    %Let $|\cX|<\infty$. For any $\cC$ over $\cX$ with $\VCD(\cC) = 1$ there exists a metric $d$ with $\cS_{\CS^d_{\min}}(\cC) \leq 2$.
    For any finite $\cC$  with $\VCD(\cC) = 1$ there exists a metric $d$ with $\cS_{\CS^d_{\min}}(\cC) \leq 2$.
\end{restatable}

The proof (see appendix) uses the fact that every $x\in\cX$ has at most one label $b\in\{0,1\}$ such that $(x,b)$ ``teaches'' a unique $C\in\cC$ in the model of recursive teaching~\cite{DoliwaFSZ14}. The instances in $\cX$ can be partitioned according to these labels. This partitioning is used to define a metric that places instances in the same part closer to one another than instances in different parts. Now a learner choosing one specific query from each of the two parts will receive enough information to identify the target. %Details are given in the supplementary material.

\subsection{Lower Bounds Using the Hamming Distance}

Let DL$_m$ be the class of 1-decision lists over the Boolean
variables $v_1,\ldots,v_m$. Each list is of shape
\begin{equation} \label{eq:decision-list}
\cL = [(\ell_1,b_1),\ldots,(\ell_z,b_z),b_{z+1}] \enspace ,
\end{equation}
where $z \ge 0$, $\ell_i \in \{v_1,\bar v_1,\ldots,v_m,\bar v_m\}$
and $b_i \in \{0,1\}$. We assume that each variable occurs at
most once in $\cL$ (negated or not negated). We say that
a point $\vec{a} \in B_m$ is \emph{absorbed by item $k$ of $\cL$}
if the literal $\ell_k$ applied to $\vec{a}$ evaluates to $1$ while the
literals $\ell_1,\ldots,\ell_{k-1}$ applied to $\vec{a}$ evaluate
to $0$. The list $\cL$ represents the following Boolean function $f_\cL$: (i) If $\vec{a}$ is absorbed by the $k$'th item of $\cL$,
then $f_\cL(\vec{a}) = b_k$. (ii)
If $\vec{a}$ is absorbed by none of the $z$ items of $\cL$,
then $f_\cL(\vec{a}) = b_{z+1}$.

% 

%  We define

% and observe that the following holds:

The following lemma observes a structural property of a concept class $\cC$ which allows us to lower-bound the sample complexity in the minimum-distance model (with the Hamming distance) using the membership query complexity of a certain subclass of $\cC$.

\begin{restatable}{lemma}{lemSubclass} \label{lem:min-dist-to-subC-MQ}
Suppose that $\cC$ is a concept class over $B_m$, and
$\cC'$ is a subclass of $\cC$ with the properties \\
\textbf{(P1)}
For each $\vec{a} \in B_m$ with $a_m = y_m$ and each $C \in \cC'$,
we have that $C(\vec{a}) = 1$.\\
\textbf{(P2)}
For each $\vec{a}$ with $a_m= y'_m$ and $a_{m-1}=y_{m-1}$ and each $C \in \cC'$,
we have that $C(\vec{a}) = 0$.\\
% In other words, each Boolean vector ending $1$ is a positive example for each $C \in \cC'$ while each Boolean vector ending $10$ is a negative example for each $C \in \cC'$.
For $i\in\{m,m-1\}$, here $y_i \in \{0, 1\}$ is fixed, $y'_i = 1 - y_i$, and $\cC''$ is the set of all concepts of shape
% ~(\ref{eq:hard-subclass}).
\begin{equation*} %\label{eq:hard-subclass}
%\cC'' = \{
(a_1,\ldots,a_{m-2}) \mapsto C(a_1,\ldots,a_{m-2}, y'_{m-1}, y'_{m})
\mbox{ for } C \in \cC'\,.
%\}
\end{equation*}
Then $\cS_{\CS^d_{\min}}(\cC) \ge\cS_{\CS^d_{\min}}(\cC') \geq \cS[\MQ](\cC'')$, where $d$ is the Hamming distance.
%Then the number of contrast-oracle queries needed to learn $\cC$ (or $\cC'$) in the minimum-distance model is lower-bounded by the number of membership queries needed to learn $\cC''$.
\end{restatable}

\emph{Proof.} \emph{(Sketch.)}
Suppose $L$ learns $\cC'$ from $q$ queries to a contrast oracle in the minimum distance
model. We show that $L$ can be
transformed into $L'$ which learns $\cC''$ from at most $q$ queries to a membership oracle. To this end, let $C''$ be the target
concept in $\cC''$ and let $C'$ with $C'(\vec{a}y'_{m-1}y'_{m}) = C''(\vec{a})$ be
the corresponding concept in $\cC'$. Note that $L$ has not uniquely
identified $C'$ in $\cC'$ as long as the subfunction $C''$ is not uniquely
identified in $\cC''$. $L'$ can therefore identify $C''$ in $\cC''$
by maintaining a simulation of $L$ until $L$ has identified $C'$
in $\cC$. %In order to explain how $L'$ simulates the contrast oracle, we proceed by case distinction; details are given in the appendix.
Details are given in the appendix.
\iffalse
\begin{itemize}
\item
If $L$ chooses a query point of the form $\vec{a}y'_{m-1}y'_m$, then $L'$ chooses
the query point $\vec{a}$ and receives the label $b := C''(\vec{a})$
from its membership oracle. Then $L'$ returns $b$ and the point $\vec{a'}$
to $L$ where $\vec{a'} =\vec{a}y'_{m-1}y_m$ if $b=0$ and $\vec{a'} = \vec{a}y_{m-1}y'_m$
if $b=1$. Note that $(b,\vec{a'})$ is among the admissible answers of
the contrast oracle.
\item
If $L$ chooses a query point of the form $\vec{a}y_{m-1}y'_m$, then $L'$
returns the label $0$ and the extra-point $\vec{a}y_{m-1}y_m$. Again, this is
among the admissible answers of the contrast oracle.
\item
If $L$ chooses a query point of the form $\vec{a}y_{m-1}y_m$, then $L'$
returns the label $1$ and an the extra-point $\vec{a}y_{m-1}y'_m$. Again,
this is among the admissible answers of the contrast oracle.
\item
If $L$ chooses a query-point of the form $\vec{a}y'_{m-1}y_m$, then $L'$
chooses the query point $\vec{a}$ and receives the label $b = C''(\vec{a})$
from its membership oracle. Then $L'$ returns the label $b$ and
the point $\vec{a'}$ to $L$ where $\vec{a'} = \vec{a}y'_{m-1}y'_m$ if $b=0$
and $\vec{a'} = \vec{a}y_{m-1}y'_m$ if $b=1$. Again, this is among the admissible
answers of the contrast oracle.
\end{itemize}
Clearly $L''$ can maintain this simulation until it reaches exact identification
of the target concept. Moreover, the number of query point chosen by $L'$
does not exceed the number of query points chosen by $L$.
\fi
\hfill$\Box$

This lemma can be used, for example, to prove $\cS_{\CS^d_{\min}}(\DL_m)\ge 2^{m-2}-1$, where $d$ is the Hamming distance. Our proof of this statement, however, establishes an even stronger result. To formulate this stronger result, we first need to introduce some notation.

\iffalse
\emph{Proof.}
Let $\cC = \mbox{DL}_m$. Moreover, let $\cC' \subseteq \cC$ be a subset of DL$_m$ of decision lists of the form
\[
\cL = [(v_m,1) , (v_{m-1},0) ,
\underbrace{(\ell_1,b_1) , \ldots (\ell_z,b_z) , b_{z+1}}_{= \cL' \in \mbox{\small DL}_{m-2}}] \in \mbox{DL}_m
\enspace ,
\]
where $z \ge 0$, $\ell_i \in \{v_1,\bar v_1,\ldots,v_{m-2},\bar v_{m-2}\}$
and $b_i \in \{0,1\}$. $\cC'$ follows the property (P1) and (P2) with $y_{m-1} = y_{m} = 1$ in Lemma~\ref{lem:min-dist-to-subC-MQ}. Then $\cC'' = \mbox{DL}_{m-2}$.  $\MONOMIALS^{m-2}$
is a subclass of DL$_{m-2}$ and $2^{m-2}-1$ membership queries are needed
for learning $\MONOMIALS^{m-2}$.
\hfill$\Box$
\fi

Consider a decision list $\cL$ of the form~(\ref{eq:decision-list}).
We say that $\cL$ has $k$ label alternations if the number of distinct
indices $i \in [z]$ such that $b_{i+1} \neq b_i$ equals $k$.
Let DL$_m^k$ be the subclass of DL$_m$ which contains all decision
lists with at most $k$ label alternations. It is well known that (i) DL$_m^0$ contains only the constant-1 and the constant-0 function, and (ii) DL$_m^1 = \MONOMIALS^m \cup \CLAUSES^m$.
%Now an inspection of the proof of Corollary~\ref{cor1:decision-list-lb} reveals an actually stronger result:
Now the proof of the following claim builds on Lemma~\ref{lem:min-dist-to-subC-MQ}; see the appendix for details.

%\begin{restatable}{corollary}{cordecisionlistPP} \label{cor1:decision-list-lb}
%$\cS_{\CS^d_{\min}}(\DL_m)\ge 2^{m-2}-1$, where $d$ is the Hamming distance. 
%At least $2^{m-2}-1$ contrast-oracle queries are needed for learning DL$_m$ in the minimum-distance model.
%\end{restatable}

\begin{restatable}{corollary}{cordecisionlistPP}\label{cor2:decision-list-lb}
$\cS_{\CS^d_{\min}}(\DL^2_m)\ge 2^{m-2}-1$, where $d$ is the Hamming distance.
%At least $2^{m-2}-1$ contrast-oracle queries are needed for learning DL$_m^2$ in the minimum-distance model.
\end{restatable}

Interestingly, this bound on $\cS_{\CS^{d}_{\min}}(\DL_m)$ (using Hamming distance) is not an effect of our metric-independent lower bound in terms of $\SD$ (Thm.~\ref{thm:sd}). As seen below,  $\cS_{\CS^{d}_{\min}}(\DL_m)$ 
asymptotically exceeds $\operatorname{SD}(\DL_m)$.
Thus, the lower bound from Thm.~\ref{thm:sd}
is not always asymptotically tight. %, not even for natural concept classes combined with natural metrics. 

For $\DL_m$, we saw that the sample complexity in the minimum Hamming distance model is exponential in $m$. By contrast, $\SD$ is at most quadratic in $m$. To prove this, let a \emph{block}\/ in  $[(\ell_1,b_1),$ $\ldots,(\ell_z,b_z),b_{z+1}]$ be any maximal substring $(\ell_i,b_i),
\ldots,(\ell_{i+j},b_{i+j})$ with $b_i=\ldots = b_{i+j}$.

%Let $\mathcal{L}=[(\ell_1,b_1),\ldots,(\ell_z,b_z),b_{z+1}]$ be a 1-decision list. A \emph{block}\/ in $\mathcal{L}$ is any maximal substring of the form $(\ell_i,b_i),(\ell_{i+1},b_{i+1}),\ldots,(\ell_{i+j},b_{i+j})$ such that $b_i=b_{i+1}=\ldots = b_{i+j}$. %The number of alternations in $\mathcal{L}$ is defined to be the number of blocks minus 1.

\begin{restatable}{theorem}{thmsddl} \label{thm:DL-SD}
There exists a self-directed learner for $\DL_m$ that makes, on any  target list $C^*\in\DL_m$, at most $4km$ mistakes where $k$ is the number of blocks in  $C^*$.
\end{restatable}
\emph{Proof.} \emph{(Sketch.)}
    For each of the $2m$ literals, 2 queries suffice to determine whether it occurs in the first block of the target list. %Proceeding i
    Iteratively, % with the remaining blocks 
    one consumes $\le 4m$ queries per block; see the appendix.
\hfill$\Box$

We now present a second example of a natural class of Boolean functions for which $\SD$ is asymptotically smaller than 
$\cS_{\CS^{d}_{\min}}(\cC)$, where $d$ is the Hamming distance.

\begin{definition}
    Fix $s,z,m\in\mathbb{N}$. An $s$-term $z$-MDNF of size $m$ is a function $f:\{0, 1\}^m \rightarrow \{0, 1\}$ of the form $f(v_1, v_2, \cdots, v_m) = M_1 \vee M_2 \vee \cdots \vee M_s$, where each $M_i$ is a monotone monomial with at most $z$ literals. We use $\MDNF{m}{s}{z}$ to refer to the class of all $s$-term $z$-MDNF of size $m$.
\end{definition}

\iffalse
\begin{theorem}[largely due to \cite{GoldmanS94,abasi2014exact}]\label{thm:MDNF-min-dist-MQ}
    Let $d$ be the Hamming distance. Then $\SD(\MDNF{m}{s}{z})=s$, while $\cS_{\CS^d_{\min}}\ge\mathcal{S}[\MQ](\MDNF{m-2}{s-1}{z-1})$, which is at least
    $$%\cS_{\CS^d_{\min}}(\MDNF{m}{s}{z})
         %\geq \mathcal{S}[\MQ](\MDNF{m-2}{s-1}{z-1})\ge  
         (z - 1)(s - 1)\log (m - 2) + \begin{cases}\left(\frac{z - 1}{s - 1}\right)^{s - 2} & z > s\,,\\
    \left(\frac{2s - 2}{z - 1}\right)^{(z - 1)/2} & z \leq s\,.
    \\
    \end{cases}$$
\end{theorem}
\fi
\begin{theorem}[largely due to \cite{GoldmanS94,abasi2014exact}]\label{thm:MDNF-min-dist-MQ}
    Let $d$ be the Hamming distance. Then $\SD(\MDNF{m}{s}{z})=s$, while $\cS_{\CS^d_{\min}}\ge\mathcal{S}[\MQ](\MDNF{m-2}{s-1}{z-1})$, which is at least
    $$
         (z - 1)(s - 1)\log (m - 2) + \alpha\,,
    $$
   where $\alpha=\left(\frac{z - 1}{s - 1}\right)^{s - 2}$ if $ z > s$, and $\alpha=
\left(\frac{2s - 2}{z - 1}\right)^{(z - 1)/2}$ if $z \leq s$.
\end{theorem}
\emph{Proof.}
See \cite{GoldmanS94} for $\SD(\MDNF{m}{s}{z})=s$. \cite{abasi2014exact} showed the lower bound on $\mathcal{S}[\MQ](\MDNF{m-2}{s-1}{z-1})$.

Let $\cC^{\prime}$ be the subclass of $\MDNF{m}{s}{z}$ with concepts of the form $M_1 \vee \ldots \vee M_{s-1}\vee v_m$ where $v_{m-1} \in M_i$ for $1\le i\le s-1$. Then $\cC'$ satisfies properties (P1) and (P2) of Lemma~\ref{lem:min-dist-to-subC-MQ} with $y_m = 1$, and $y_{m-1} = 0$. Then $\cC'' = \mathrm{MDNF}_{m-2, z-1, s-1}$, and the result follows from Lemma~\ref{lem:min-dist-to-subC-MQ}.
\hfill$\Box$

\iffalse
\begin{remark}
    \cite{abasi2014exact} has proved that 
    $$\mathcal{S}[\MQ](\MDNF{n}{s}{r}) \geq  rs\log n + \begin{cases}\left(\frac{r}{s}\right)^{s - 1} & r > s\\
    \left(\frac{2s}{r}\right)^{r/2} & s \leq r
    \\
    \end{cases}$$
\end{remark}
\fi

%\section{Dynamic Metrics}

Theorem~\ref{thm:MDNF-min-dist-MQ} reveals a gap between $\SD$ and the sample complexity of the minimum distance model under the Hamming distance. While we do not know of any ``natural'' function $d:\mathcal{X}\times\mathcal{X}\rightarrow\mathbb{R}^{\ge 0}$ with which to close this gap, it can be closed if we allow the learner and the contrast oracle to share a new function $d$ after every query. %Important is that the learner knows the sequence of distance functions used by the oracle. 
By $\cS_{\CS^{(d^t)}_{\min}}$, we denote the sample complexity resulting from interaction sequences at which, in step $t$, the set $\CS$ is defined using the distance function $d^t$.
The same proof as for Theorem~\ref {thm:sd} shows that our lower bound in terms of $\SD$ (and thus the lack of an upper bound in terms of $\VCD$, cf. Remark~\ref{rem:vcd}) persists despite this strengthening of the model.

\begin{theorem}
    Fix $(d_t)_{t = 1}^{\infty}$ and $\cC$ over a finite domain $\cX$. Then  $\cS_{\CS^{(d^t)}_{\min}}(\cC) \geq \lceil\operatorname{SD}(\cC)/2\rceil$.
\end{theorem}

However, %as we will see below, 
the dynamic distance function approach helps to overcome obstacles noted in Theorem~\ref{thm:MDNF-min-dist-MQ}. One natural way to define dynamically changing functions $d^t$ is to make them dependent on the version space.
%For any $\cC' \subset \cC$ define $d_{\cC'}(x, x') := {|\{C \in \cC': C(x) \neq C(x')\}|}/{|\cC'|}$. 
Defining $d_{\cC'}(x, x') := {|\{C \in \cC': C(x) \neq C(x')\}|}/{|\cC'|}$ for any $\cC' \subset \cC$, and recalling that $\cC_t$ refers to the version space obtained from $\cC$ after $t$ queries, we get (see  appendix for a proof):%, Define \emph{version space metric} as this distance on the current version space $d^t_{VS} := d_{\cC_t}$.

\iffalse
BEGIN FOR APPENDIX
\begin{lemma}\label{lem:mon-MDNF}
    Let $\vec x, \vec y, \vec z \in \{0, 1\}^n$ such that the 1s in $\vec x$ is subset of the 1s in $\vec y$, which is subset of the 1s in $\vec z$. Then for any $\cC' \subset \MDNF{n}{s}{r}$ we have $d_{\cC'}(\vec x, \vec y) \leq d_{\cC'}(\vec x, \vec z)$.
\end{lemma}
\emph{Proof.}
    Consider any $C \in \MDNF{n}{s}{r}$. Since every MDNF is monotonic, if $C(\vec x) = 1$ then we have $C(\vec y) = 1, C(\vec z) = 1$. Moreover, similarly if $C(\vec x) = 0$ and $C(\vec y) = 1$ then $C(\vec z) = 1$ as well. This completes the proof. 
\hfill$\Box$
END FOR APPENDIX
\fi

\begin{restatable}{theorem}{thmmdnf}\label{thm:mdnf2}
   %Then \YC{drop ``Then''?} 
   $\cS_{CS^{(d_{\cC_t})}_{\min}}(\MDNF{m}{s}{z}) \leq s = \operatorname{SD}(\MDNF{m}{s}{z})$, where $\cC_t=(\MDNF{m}{s}{z})_t$ for any $t$.
\end{restatable}

\section{Conclusions}

We proposed and analyzed a  generic  framework for active learning with contrastive examples. In studying the sample complexity in two of its instantiations (the minimum distance and the proximity model), we observed interesting connections to other models of query learning, and most notably to self-directed learning, which also led to a new result on self-directed learning of decision lists (Theorem~\ref{thm:DL-SD}).  Our definition allows modelling contrastive learning with a multitude of contrast set rules, and it can easily be extended to encompass passive learners receiving contrastive examples.

Our framework allows the learner to reason with perfect knowledge about the choice of the contrast set. While this is unrealistic in some practical situations, it has applications in formal methods, %specifically in \emph{counter-example guided synthesis}\/ \cite{AbateDKKP18}, 
where the contrast oracle is a computer program that could be simulated by the learner in order to refute a hypothesis (a concept $C$ can be excluded if the oracle does not provide the same counterexamples when $C$ is the target). %Adaptations to address this are conceivable, but will likely affect our results on sample complexity. 
%Nevertheless, o
Moreover, the learner's knowledge of the mapping $\CS$ implicitly connects our framework to \emph{self-supervised learning}. Consider, for example, passive learning under the minimum distance model. Here the learner is passively sampling examples $(x,y)$, which are supplemented with contrastive examples from $\CS^d_{\min}$. Following our model, the learner knows that the contrastive example $x'$ to $x$ is the closest one to $x$ that has a label different from $y$. In practice, this can be implemented as a self-supervised mechanism in which the learner internally labels with $y$ all data points that are strictly closer to $x$ than $x'$.
In future work, the condition that the learner has perfect knowledge of the set $\CS$ can be softened so as to encompass more learning settings.% in our framework.
%Our framework can serve as a useful fundamental tool for the formal study of learning from contrastive examples.

% \bibliographystyle{apalike}
%\bibliography{ref}

%%%%%%%%%%%%%%%%%%%%%%%%%%%%%%%%%%%%%%%%%%%%%%%%%%%%%%%%%%%%

\bibliographystyle{plain}
\bibliography{ref}

%%%%%%%%%%%%%%%%%%%%%%%%%%%%%%%%%%%%%%%%%%%%%%%%%%%%%%%%%%%%
\iffalse
\appendix

\section{Technical Appendices and Supplementary Material}
Technical appendices with additional results, figures, graphs and proofs may be submitted with the paper submission before the full submission deadline (see above), or as a separate PDF in the ZIP file below before the supplementary material deadline. There is no page limit for the technical appendices.
\fi
%%%%%%%%%%%%%%%%%%%%%%%%%%%%%%%%%%%%%%%%%%%%%%%%%%%%%%%%%%%%

\appendix

\section{Appendix}

This appendix provides the proofs omitted from the main paper.

%\setcounter{theorem}{1}
\iffalse
%\begin{restatable}{proposition}{propinjective}
Let $\mathcal{C}$ be a countable concept class over a countable $\mathcal{X}$. Let $T:\mathcal{C}\rightarrow 2^\mathcal{X}$ be any injective function that maps every concept in $\mathcal{C}$ to a finite set of instances. Then: 
\begin{enumerate}
\item There is some $\CS$ with $\mathcal{S}_{\CS}(\mathcal{C})\le\sup_{C\in\mathcal{C}}|T(C)|+1$. 
\item If $T(C)\not\subseteq T(C')$ for $C\ne C'$, then there is some $\CS$ such that $\mathcal{S}_{\CS}(\mathcal{C})\le\sup_{C\in\mathcal{C}}|T(C)|$.
\end{enumerate}
In particular, if $\cX$ is finite, then there is some $\CS$ such that $\mathcal{S}_{\CS}(\mathcal{C})\le 1+\min\{k\mid\sum_{i=0}^k\binom{|\cX|}{i}\ge|\cC|\}$. If $\mathcal{X}$ is countably infinite, then there is some $\CS$ such that $\mathcal{S}_{\CS}(\mathcal{C})=1$. 
%\end{restatable}
\fi

\propinjective*

\emph{Proof.}
Let $T:\mathcal{C}\rightarrow\mathcal{X}$ be such that $T(C)\neq T(C')$ whenever $C\ne C'$. Let $x_1,x_2,\ldots$ be a fixed repetition-free enumeration of all the elements in $\mathcal{X}$. 

To prove statement 1, define $\CS$ by
$\CS(x_i,C) = \{x_{j'}\}\ \mbox{ for }\ j' = \min\{j\mid j \ge i,\ x_j \in T(C)\}$.
Now an active learner operates as follows. Initially, it sets $n_1=1$ and starts with iteration 1. In iteration $i$, it asks a query for $x_{n_i}$. If it receives $x'_i=x_j$ as a contrastive example, then note that $j\ge n_i$. The learner will then set $n_{i+1}=j+1$ and proceed to iteration $i+1$. 

Upon its $i$th query, the learner will find out the label for $x_{n_i}$ (which is irrelevant) and the $i$th element of $T(C)$ in the enumeration $x_1,x_2,\ldots$. After $|T(C)|$ queries, the learner has seen all elements from $T(C)$, and will receive a dummy response for its next query. From its responses, it has now inferred that its target is a concept that is mapped to $T(C)$ by mapping $T$. By injectivity of $T$, this means that the learner has uniquely identified $C$. This proves statement 1. Since an injective mapping $T$ using all subsets of $\cX$ of size at most $k^*$ can encode $\sum_{i=0}^{k^*}\binom{|\cX|}{i}$ concepts, such $T$ would yield $\mathcal{S}_{\CS}(\mathcal{C})\le 1+\min\{k\mid\sum_{i=0}^k\binom{|\cX|}{i}\ge|\cC|\}$.

For statement 2, note that, if $T(C)\not\subseteq T(C')$ for distinct $C,C'\in\mathcal{C}$, then the learner can infer $C$ correctly after having seen all elements in $T(C)$. With this result, for countably infinite $\cX$, any mapping $T$ identifying each concept with a different instance in $\cX$, results in $\mathcal{S}_{\CS}(\mathcal{C})=1$.
\hfill$\Box$

\remprox*
\emph{Proof.} 
   The second inequality holds by definition. For the first inequality, consider any $r > 0$ and $(x,C) \in \cX \times \cC$. Due to Remark~\ref{rem:completeness}, $\CS^d_{\min}(x, C) = \emptyset$ iff there is no $x'$ with $C(x')\ne C(x)$. Moreover, $\CS^d_{\prox}(x, r, C) = \emptyset$ iff there are no points with different label converging to a point at distance at most $r$ from $x$. Therefore, either  $\CS^d_{\prox}(x, r, C) = \emptyset$, or $\CS^d_{\prox}(x, r, C) \supseteq \CS^d_{\min}(x, C) \neq \emptyset$. In the former case, the contrastive oracle in the minimum distance model also conveys that there are no points with different label converging to a point at distance at most $r$ from $x$. Consequently, the contrastive oracle in the proximity model does not provide extra information. In the later case, any contrastive example given by the contrastive oracle in the minimum distance model can be also given by the contrastive oracle in the proximity model. Again, the contrastive oracle in the proximity model does not provide extra information.
   % he same reasoning as for Remark~\ref{rem:cs} applies.
\hfill$\Box$

\exmmonfirst*
\iffalse
\begin{restatable}{example}{exmmonfirst}\label{exmp:monomials1}
Let $\mathcal{X}=\{0,1\}^m$. Let $\mathcal{C}^m_{\pmon}$ consist of all monotone monomials, i.e., logical formulas of the form $v_{i_1}\wedge \ldots\wedge v_{i_k}$ for some pairwise distinct $i_1,\ldots,i_k$ and some $k\in\{0,\ldots,m\}$. The concept associated with such a formula contains the boolean vector $(b_1,\ldots,b_m)$ iff $b_{i_1}=\ldots =b_{i_k}=1$ (where the empty monomial is the constant 1 function). 
Let $d$ be the Hamming distance. 
% and let \[\CS(x,C^*)=\arg\min_{x'\in\mathcal{X},C^*(x)\ne C^*(x')}d(x,x')\,.\]
Then:  %the following statements hold.
\iffalse
$$
(1)\ \mathcal{S}[\MQ](\mathcal{C}^m_{\pmon}) = m; \ \ (2)\ \mathcal{S}_{\CS^d_{\min}}(\mathcal{C}^m_{\pmon})=1;\ \ (3)\ \mathcal{S}_{\CS^d_{\prox}}(\mathcal{C}^m_{\pmon})=\Theta(\log m)\,.
$$
\fi
\begin{enumerate}
    \item  $\mathcal{S}[\MQ](\mathcal{C}^m_{\pmon}) = m$.
    \item $\mathcal{S}_{\CS^d_{\min}}(\mathcal{C}^m_{\pmon})=1$.
    \item  $\mathcal{S}_{\CS^d_{\prox}}(\mathcal{C}^m_{\pmon})=\Theta(\log m)$.
\end{enumerate}
%\end{restatable}
\fi

\emph{Proof.} It remains to prove the lower bound from Claim (3).

\iffalse
\emph{Upper bound:} 
 Let the target concept be $v_{i_1} \wedge\ldots\wedge v_{i_k}$.
Query $(\vec{0},k)$ would force the oracle to return a Boolean vector with
ones exactly in the positions $i_1,\ldots,i_k$. $L$ can determine $k$ by
means of binary search. The query $(\vec{0},m/2)$ halves the search space
for $k$: if a contrastive example is returned by the oracle, then $k \le m/2$;
if $\omega$ is returned, then $k > m/2$.
\iffalse
 A learner first queries $\vec 0$ with $r_1 = \lfloor \frac{m}{2} \rfloor$. Let the target concept be $v_{i_1}\wedge \ldots\wedge v_{i_k}$. Since every vector with $b_{i_1}=\ldots =b_{i_k}=1$ has Hamming distance $\ge k$ with $\vec 0$, a contrastive example exists iff $k \le r_1$. If $k > r_1$, querying $\vec 0$ with $r_2 = \lfloor\frac{r_1 + m}{2} \rfloor$ will tell the learner whether or not $k > r_2$. If $k \leq r_1$, querying $\vec 0$ with $r_2 = \lfloor\frac{r_1}{2} \rfloor$ tells the learner whether or not $k \in (r_2, r_1]$. Thus, binary search yields $r_{t} = k$ for $t \leq \log m$. When querying $\vec 0$ with radius $k$, the target vector is the only vector whose Hamming distance from $\vec 0$ is at most $k$. Thus $\mathcal{S}_{\CS^d_{\prox}}(\mathcal{C}^m_{\pmon}) \leq \log m$.
\fi
\fi
 
 Let $(\vec{x},r)$ be the first query point of the learner. Let $w$ be the
Hamming weight of $\vec{x}$. The logarithmic lower bound is now obtained
by an adversary argument.

If $w > m/3$, then the oracle returns $(\vec{x},1)$
and $(\vec{x'},0)$ where $\vec{x'}$ is obtained from $\vec{x}$
by flipping one of the $1$-entries in $\vec{x}$ to $0$. Note that,
if $j_1,\ldots,j_{w-1}$ denote the positions of the $1$-entries
in $\vec{x'}$, then the index set associated with the variables in $C^*$
could be any subset
of $\{j_1,\ldots,j_{w-1}\}$.
 
If $r \geq m/3$ and $w \le m/3$, then we may assume without loss of generality that $r \leq m-w$.
The oracle returns $(\vec{x},0)$
and $(\vec{x'},1)$ where $\vec{x'}$ is obtained by adding $r$ ones to $\vec{x}$,
Note that, if $j_1,\ldots,j_r$ denote the positions of the additional ones,
then the index set associated with the variables in $C^*$ could be any
subset of $\{j_1,\ldots,j_r\}$.
 
If $r < m/3$ and $w \le m/3$, then the oracle returns $(\vec{x},0)$
and $\omega$. %Fix an index set $I \subset [m]$ of size $m-r-w$ such that $x_i = 0$ for each $i \in I$. % and note that the index set associated with the variables in $C^*$ could be any subset of $I$.
Let $J$ with $|J| = w$ be the set of indices $j$ such that $x_j = 1$.
Fix a set $K \subseteq [m]$ such that $|K| = r+1$ and $K \cap J = \emptyset$.
Let $I = [m] \sm (J \cup K)$ be the set of the remaining indices in $[m]$.
Let us make the commitment that the variables $x_k$ with $k \in K$ are contained in $C^*$.
Even if the learner knew about this commitment, it would still have to determine the $I$-indices of the remaining variables in $C^*$. These indices could form an arbitrary subset of $I$.
 
In any case the learner can reduce the search space for the index set
of $C^*$ not more than by a factor of $3$. Hence the learner
requires $\Omega(\log m)$ queries for the exact identification of $C^*$.
\hfill$\Box$

\exmmon*
\iffalse
 Let $\mathcal{X}'=\{0,1\}^{m+1}$. Define a concept class $\mathcal{C}'_{\pmon}$ over $\mathcal{X}'$ from $\mathcal{C}^m_{\pmon}$ as follows. Each concept $C\in\mathcal{C}^m_{\pmon}$ is extended to a concept $C'\in\mathcal{C}'_{\pmon}$ via $C'(b_1,\ldots,b_m,0)=C(b_1,\ldots,b_m)$ and $C'(b_1,\ldots,b_m,1)=1-C(b_1,\ldots,b_m)$.
 No other concepts are contained in $\mathcal{C}'_{\pmon}$. 
 Let $d$ be the Hamming distance.
 % and let \[\CS(x,C^*)=\arg\min_{x'\in\mathcal{X}',C^*(x)\ne C^*(x')}d(x,x')\,.\]
 Then $\mathcal{S}[\MQ](\mathcal{C}'_{\pmon})=\mathcal{S}_{\CS^d_{\min}}(\mathcal{C}'_{\pmon})=\mathcal{S}_{\CS^d_{\prox}}(\mathcal{C}'_{\pmon})=m$.
 \fi
 %Then the following statements hold.
%\begin{enumerate}
 %   \item $\MQ(\mathcal{C}'_{\pmon})=m$.
 %   \item $\mathcal{S}_{\CS^d_{\min}}(\mathcal{C}'_{\pmon})=m$.
%    \item  $\mathcal{S}_{\CS^d_{\prox}}(\mathcal{C}'_{\pmon})=m$.
%\end{enumerate}
%\end{restatable}

 \emph{Proof.}
 A membership query learner using $m$ queries will work as described in the proof of Example~\ref{exmp:monomials1}, yet will leave the $(m+1)$st component of every queried vector equal to 0. Thus, the learner will identify which vectors of the form $(b_1,\ldots,b_m,0)$ belong to the target concept. The latter immediately implies which vectors of the form $(b_1,\ldots,b_m,1)$ belong to the target concept. Again, by a standard information-theoretic argument, fewer membership queries will not suffice in the worst case. Thus $\mathcal{S}[\MQ](\mathcal{C}'_{\pmon})=m$.
 
 Upon asking any query $(b_1,\ldots,b_m,b_{m+1})$, a learner in the minimum distance model will receive the correct answer, plus the contrastive example $(b_1,\ldots,b_m,1-b_{m+1})$. This contrastive example does not provide any additional information to the learner. Hence, the learner has to ask as many queries as an $\MQ$-learner, i.e., $\mathcal{S}_{\CS^d_{\min}}(\mathcal{C}'_{\pmon})=m$.
 
  Finally, according to Remark~5, $\mathcal{S}_{\CS^d_{\min}} \leq \mathcal{S}_{\CS^d_{\prox}} \leq \mathcal{S}[\MQ]$. Thus, $\mathcal{S}_{\CS^d_{\prox}}(\mathcal{C}'_{\pmon})=m$.
\hfill$\Box$

\begin{remark}
    Example~\ref{exmp:monomials2} can be generalized: each concept class $\mathcal{C}$ over domain $\mathcal{X}=\{0,1\}^n$ can be (redundantly) extended to a concept class $\mathcal{C}'$ over  $\mathcal{X}'=\{0,1\}^{n+1}$ that renders contrastive examples in the minimum distance model useless for learning concepts in $\mathcal{C}'$, under the Hamming distance.
\end{remark}

\coralternation*
\iffalse
\setcounter{theorem}{6}

\begin{restatable}{theorem}{coralternation}\label{cor:alternation1}
$\mathcal{S}_{\CS^d_{\min}}(\MONOMIALS^m \cup \CLAUSES^m)=2$, where $d$ is the Hamming distance.
%The class $\MONOMIALS^m \cup \CLAUSES^m$ can be learned from a contrast oracle in the minimum-distance model at the expense of $2$ queries.
\end{restatable}
\fi

To prove this result, we begin with a straightforward extension of Example~\ref{exmp:monomials1}.

\setcounter{theorem}{22}
\begin{restatable}{lemma}
{propmonomials}
\label{lem:monomials}
$\mathcal{S}_{\CS^d_{\min}}(\MONOMIALS^m)=2$, where $d$ is the Hamming distance.
%$\MONOMIALS^m$ can be learned from a contrast oracle in the minimum-distance model at the expense of $2$ queries.
\end{restatable}

\emph{Proof.}
    The proof is analogous to that of Example~\ref{exmp:monomials1}, just here the learner asks two queries, one for $\vec{0}$ and one for $\vec{1}$. The two contrastive examples then determine the target monomial, by a similar reasoning as for Example~\ref{exmp:monomials1}.
\hfill$\Box$

By duality\footnote{The
mapping $f(\vec{v}) \mapsto \bar f(\ol{\vec{v}})$ transforms the monomial
$\bigwedge_{i \in I}\bar{v_i} \wedge \bigwedge_{j \in J}v_j$ into the clause
$\bigvee_{i \in I}\bar v_i \vee \bigvee_{j \in J}v_j$, and vice versa.
The same mapping can be used to dualize the proof of Lemma~\ref{lem:monomials}
in the sense that, within this proof, every classification label is negated
and every point from $B_m$ is componentwise negated. This concerns the query
points chosen by the learner as well as the contrastive examples returned by the oracle.}, Lemma~\ref{lem:monomials} implies the following
result:

\begin{corollary} \label{cor:clauses}
$\mathcal{S}_{\CS^d_{\min}}(\CLAUSES^m)=2$, where $d$ is the Hamming distance.
%$\CLAUSES^m$ can be learned from a contrast oracle in the minimum-distance model at the expense of $2$ queries.
\end{corollary}

We can now proceed with the proof of Theorem~\ref{cor:alternation1}.

\emph{Proof.} \emph{(of Theorem~\ref{cor:alternation1}.)}
Let $L_\wedge$ be the learner of $\MONOMIALS^m$ as it is described
in the proof of Lemma~\ref{lem:monomials}. Let $L_\vee$ be the
learner of $\CLAUSES^m$ which is obtained from $L_\wedge$ by
dualization. It suffices to describe a learner $L$ who maintains
simulations of $L_\wedge$ and $L_\vee$ and \emph{either} comes in both
simulations to the same conclusion about the target
concept\footnote{This happens when the target concept is a literal,
i.e., both a monomial of length $1$ and a clause of length $1$.}
\emph{or} realizes an inconsistency in one of the simulations,
which can then be aborted. Details follow. \\
If the target concept is representable as a literal, then $L$ will
not run into problems because both simulations lead to the same
(correct) result. From now we assume that the target concept
cannot be represented by a literal. Remember that both learners,
$L_\wedge$ and $L_\vee$, choose the two query points $\vec{1}$
and $\vec{0}$.  Let us first consider the case that the target function is the 1-constant (= empty monomial).
Then the oracle returns the label $1$ for both query points
along with an error message that there is no point of opposite
label in $B_m$. After having received this error message, $L$ can
abort both simulations because it has identified the $1$-constant
as the target function. The reasoning for the constant-0 function
is analogous. \\
Let us now consider the case that the target function can be
represented by a monomial $M^*$ of length at least $2$ resp.~by
a corresponding decision list. If $M^*$ contains at least one negated
and at least one not negated variable, then the oracle returns label $0$
for both query points plus an extra-point of opposite label.
If the target function were a clause, then the case of label $0$
for both query points can happen only if the target function would
be the 0-constant (so that no extra-point of opposite label
could be returned by the oracle). Hence $L$ may abort the simulation
of $L_\vee$ and identify the target function via $L_\wedge$.
Let us now assume that $M^*$ does not contain negated variables
so that $M^*$ is of the form $\bigwedge_{j \in J}v_j$
for some $J \seq [m]$ with $|J| \ge 2$. Then, upon query point $\vec{0}$,
the oracle returns label $0$ plus the extra-point $\vec{1}_J$.
The learner $L_\vee$ concludes from seeing label $0$ that the target
clause does not contain a negated variable. But $L_\vee$ would expect
to see an extra-point with only a single $1$-component (because a
clause with only unnegated variables can be satisfied by a single
$1$-bit in the right position). Hence $L$ can abort the simulation
of $L_\vee$ and continue with the simulation of $L_\wedge$ until $M^*$
is identified. \\
All cases that we did not discuss are dual to cases that we actually
have discussed. Thus the proof of the corollary is accomplished.
\hfill$\Box$

%\proxmonclaus*
\iffalse
\emph{Proof.}
    We prove the claim for $\MONOMIALS^m$. A learner $L$ first queries $(\mathbf 1, \lfloor\frac{m}{2}\rfloor)$. With a binary search as in the proof of Example~\ref{exmp:monomials1}.3, $L$ learns $|I|$ with $\log m$ queries. Now a query $(\mathbf 1, |I|)$ is equivalent to querying $\mathbf 1$ in the minimum distance model. Similarly, $L$ queries $(\mathbf 0, \lfloor\frac{m}{2}\rfloor)$, and finds $|J|$ with $\log m$ queries; then the query $(\mathbf 0, |J|)$ is equivalent to querying $\mathbf 0$ in the minimum distance model. Using Proposition~\ref{lem:monomials} this concludes the proof.
\hfill$\Box$
\fi

%\setcounter{theorem}{9}

\thmplusminusmq*
\iffalse
\begin{restatable}{theorem}{thmplusminusmq} \label{rem1:d0}
%In the minimum-distance model with metric $d_0$, the following holds:
$\mathcal{S}[\EX^+,\EX^-,\MQ](\cC)-2\ \le\ \cS_{\CS^{d_0}_{\min}}(\cC)\ \le\ \mathcal{S}[\EX^+,\EX^-,\MQ](\cC)$.
\end{restatable}
\fi

\emph{Proof.}
As for the second inequality, it suffices to show that each call
of one of the oracles $\EX^+,\EX^-,\MQ$ can be simulated by a single
call of the contrast oracle. This is clearly true for a call
of the $\MQ$-oracle (because the contrast oracle is even more informative).
A call of $\EX^+$-oracle can be simulated as follows:
\begin{enumerate}
\item
Choose an (arbitrary) instance $x\in\cX$ as input of the contrast
oracle and get back a pair $(b,x')$ such that $b = C^*(x) \neq C^*(x')$.
\item
If $b=1$, then return $x$ (as the desired positive example).
Otherwise return $x'$.
\end{enumerate}
A similar reasoning shows that a call of the $\EX^-$-oracle
can be simulated by a call of the contrast oracle. Hence the
second inequality is valid. \\
As for the first inequality, it suffices to
show that $q$ queries of the contrast oracle can be simulated
by $q$ calls of the membership oracle and two extra-calls
which are addressed to the oracles providing a (positive or
negative) example. For $i=1,\ldots,q$, let $x_i$ be the $i$-th
query instance  that is given as input to the contrast oracle and
let $(b_i,x'_i)$ with $b_i = C^*(x_i) \neq C^*(x'_i)$
be the pair returned by the latter. Note that $x'_i$
can be any instance with label opposite to $b_i$ because
the discrete metric assigns the same distance value, $1$,
to any such pair $(x,x')$. The above $q$ queries can therefore
be simulated as follows:
\begin{enumerate}
\item Call the oracle $\EX^-$ and obtain an instance $y_0$
such that $C^*(y_0) = 0$.
\item Call the oracle $\EX^+$ and obtain an instance $y_1$
such that $C^*(y_1) = 1$.
\item For $i=1,\ldots,q$, choose $x_i$ as input of the $\MQ$-oracle
and get back the label $b_i = C^*(x_i)$. If $b_i=0$,
then return $(0,y_1)$ as a valid answer of the contrast oracle
when the latter is called with query instance $x_i$.
If $b_i=1$, then return $(1,y_0)$ instead.
\end{enumerate}
This completes the proof.
\hfill$\Box$

\begin{table}[] 
\caption{A concept class $\cC$ with $\SD(\cC) = \VCD(\cC) = 1$, but $\cS_{\CS^d_{\min}}(\cC) \geq 2$ for any metric $d$.}
\centering
\begin{tabular}{|l|l|l|l|}
\hline
 & $x_1$ & $x_2$ & $x_3$ \\
\hline
$C_1$                & 1     & 0     & 1     \\
\hline
$C_2$                & 0     & 0     & 1     \\
\hline
$C_3$                & 0     & 1     & 1     \\
\hline
$C_4$                & 0     & 1     & 0    \\
\hline
\end{tabular}
\label{tab:vc-1-cs-2-ex}
\end{table}

\exmpvcd*
\iffalse
\begin{restatable}{example}{exmpvcd}
    Let $\cX=\{x_1,x_2,x_3\}$. Let $\cC$ consist of the concepts $\{x_1,x_3\}$, $\{x_3\}$, $\{x_2,x_3\}$, and $\{x_2\}$. Then $\cC$ satisfies $\SD(\cC) =\VCD(\cC) = 1$, but $\cS_{\CS^d_{\min}}(\cC) \geq 2$ for any $d:\cX \times \cX \rightarrow \reals^{\geq 0}$.
\end{restatable}
\fi
\emph{Proof.} The concept class is shown in tabular form in Table~\ref{tab:vc-1-cs-2-ex}.

Let $d$ be any distance function and let $i(1),i(2),i(3)$ be any permutation
of $1,2,3$. Let $E_{\{i(1),i(2)\}}$ be the event that ($L$ chooses query
point $x_{i(1)}$ and $d(x_{i(2)},x_{i(1)}) \le  d(x_{i(3)},x_{i(1)})$)
or ($L$ chooses query point $x_{i(2)}$
and $d(x_{i(1)},x_{i(2)}) \le  d(x_{i(3)},x_{i(2)})$). Note that at least
one of the events $E_{\{1,2\}}$, $E_{\{1,3\}}$ and $E_{\{2,3\}}$ must occur.
It suffices therefore to show that, in any case, the oracle has a response
resulting in a version space of size $2$. This can be achieved as follows:
\begin{itemize}
\item
In case of $E_{\{1,2\}}$, the response $\{(x_1,0) ,(x_2,1)\}$ would lead
to the version space $\{C_3,C_4\}$.\footnote{The learner can take no
advantage from knowing which of $x_1,x_2$ is in the role of the query point
and which is in the role of the contrastive example.}
\item
In case of $E_{\{1,3\}}$, the response $\{(x_1,0) , (x_3,1)\}$ would lead
to the version space $\{C_2,C_3\}$.
\item
In case of $E_{\{2,3\}}$, the response $\{(x_2,0) , (x_3,1)\}$ would
lead to the version space $\{C_1,C_2\}$.
\end{itemize}
\hfill$\Box$

%\setcounter{theorem}{13}
%\begin{restatable}{theorem}
\vcdOne*
%}\label{thm:vcd1}
    %Let $|\cX|<\infty$. For any $\cC$ over $\cX$ with $\VCD(\cC) = 1$ there exists a metric $d$ with $\cS_{\CS^d_{\min}}(\cC) \leq 2$.
    %For any finite $\cC$  with $\VCD(\cC) = 1$ there exists a metric $d$ with $\cS_{\CS^d_{\min}}(\cC) \leq 2$.
%\end{restatable}
%\vcdOne*

To prove this theorem, we observe the following useful fact, which was noted by \cite{DoliwaFSZ14}.

\setcounter{theorem}{24}

\begin{lemma}[\cite{DoliwaFSZ14}]\label{lem:vcd-1-column}
    Let $\cC$ be any concept class with $\VCD(\cC) = 1$. Then there is at least one $(x,b)\in\mathcal{X}\times\{0,1\}$ such that there exists exactly one $C\in\mathcal{C}$ with $C(x)=b$.
\end{lemma}

\emph{Proof.} \emph{(of Theorem~\ref{thm:vcd1}).}
It is well-known that every finite concept class of VCD 1 is contained in a concept class of size $|\cX|+1$ whose VCD is still 1, see, e.g., \cite{Floyd89}. Thus, we may assume that  $\mathcal{C}=|\cX|+1$. Set $n = |\cX|$. By an iterative application
of Lemma~\ref{lem:vcd-1-column}, we see that there exists an
order $C_1,\ldots,C_n,C_{n+1}$ of the concepts in $\cC$, an
order $x_1,\ldots,x_n$ of the instances in $\cX$ and a binary
sequence $b_1,\ldots,b_n$ such that, for $i=1,\ldots,n$, $C =C_i$ is the
only concept in $\cC\sm\{C_1,\ldots,C_{i-1}\}$ which satisfies
the equation $C(x_i) = b_i$. Let $i^+(1) <\ldots< i^+(r)$
and $i^-(1) <\ldots< i^-(s)$ with $r+s = n$ be given by $b_{i^+(j)} = 1$
for $j=1,\ldots,r$ and $b_{i^-(j)} = 0$ for $j=1,\ldots,s$.
In the sequel, we assume that $r,s \ge 1$.\footnote{The proof will
become simpler if $r$ or $s$ equals $0$.}
The following implications are easy to verify.
\begin{description}
\item[(I1)]
If $C(x_j) = 1-b_j$ for all $j\in[n]$, then $C = C_{n+1}$.
\item[(I2)]
If $C(x_{i^+(j)}) = 0$ for all $j \in[r]$, and $j^-$
is the smallest $j \in [s]$ with
$C(x_{i^-(j)}) = 0$, then $C = C_{i^-(j^-)}$.
\item[(I3)]
If $C(x_{i^-(j)}) = 1$ for all $j\in[s]$, and $j^+$
is the smallest $j \in [r]$ with 
$C(x_{i^+(j)}) = 1$, then $C = C_{i^+(j^+)}$.
\item[(I4)]
If (i) $j^-$ is the smallest $j \in [s]$ such that
$C(x_{i^-(j)}) = 0$, (ii) $j^+$ is the smallest $j \in [r]$
such that $C(x_{i^+(j)}) = 1$, and (iii) $i^+(j^+) > i^-(j^-)$,
then $C = C_{i^-(j^-)}$.
\item[(I5)]
If (i) $j^-$ is the smallest $j \in [s]$ such that
$C(x_{i^-(j)}) = 0$, (ii) $j^+$ is the smallest $j \in [r]$
such that $C(x_{i^+(j)}) = 1$, and (iii) $i^-(j^-) > i^+(j^+)$,
then $C = C_{i^+(j^+)}$.
\end{description}
It thus suffices to choose a metric $d$ such that the oracle
can be forced (with $2$ minimum-distance queries) to make
one of the above five implications applicable. To this end, we define
\[
d(x_i,x_j) = \left\{ \begin{array}{ll}
              |i-j| & \mbox{if $b_i = b_j$} \\
              n     & \mbox{if $b_i \neq b_j$}
             \end{array} \right . \enspace .
\]
The two query points, chosen by the learner, are $x_{i^+(1)}$
and $x_{i^-(1)}$. 
%The case where the oracle returns label $1$ for $x_{i^+(1)}$  or label $0$ for $x_{i^-(1)}$ is relatively easy to handle. 

Suppose first that the oracle returns the label $1$ for $x_{i^+(1)}$.
If $i^+(1) = 1$, then $C = C_1$ and we are done.
We may therefore assume that $i^+(1) > 1$, which implies that $i^-(1) = 1$.
If the oracle returns the label $0$ for $x_{i^-(1)} = x_1$, 
then $C = C_1$ and we are done.
We may therefore assume that the oracle returns the label $1$ for $x_1$.
If there is no contrastive example for $(x_1,1)$, then (I3) applies.
Otherwise, let $(x_j,0)$ be the contrastive example for $(x_1,1)$.
If $j < i^+(1)$, then (I4) applies. If $j > i^+(1)$,
then either (I3) (in case $d(x_1,x_j) = n$) or (I5)
(in case $d(x_1,x_j) < n$) applies.

Due to symmetry, one can reason analogously if the oracle returns the label $0$ for $x_{i^-(1)}$.

Finally, consider the case that the oracle returns the label $0$ for $x_{i^+(1)}$ and the label $1$ for $x_{i^-(1)}$.
%It is easy to see that one minimum distance
%query is enough unless, as assumed in the sequel, the oracle returns
%label $0$ for $x_{i^+(1)}$ and label $1$ for $x_{i^-(1)}$. 
It follows
that the contrastive example for $x_{i^+(1)}$ must be labeled $1$
and the contrastive example for $x_{i^-(1)}$ must be labeled $0$.
The following implications are easy to verify:
\begin{enumerate}
\item
If the contrastive example for $x_{i^+(1)}$ is an instance $x_i$ with $b_i=0$ and the contrastive example for $x_{i^-(1)}$ is an instance $x_i$ with $b_i = 1$, then (I1) applies.
\item
If the contrastive example for $x_{i^+(1)}$ is an instance $x_i$
with $b_i=0$ and the contrastive example for $x_{i^-(1)}$ is an
instance of the form $x_{i^-(j^-)}$, then (I2) applies.
\item
If the contrastive example for $x_{i^-(1)}$ is an instance $x_i$
with $b_i=1$ and the contrastive example for $x_{i^+(1)}$ is an
instance of the form $x_{i^+(j^+)}$, then (I3) applies.
\item
If (i) the contrastive example for $x_{i^+(1)}$ is an instance
of the form $x_{i^+(j^+)}$, (ii) the contrastive example
for $x_{i^-(1)}$ is an instance of the form $x_{i^-(j^-)}$,
and (iii) $i^+(j^+) > i^-(j^-)$, then (I4) applies.
\item
If (i) the contrastive example for $x_{i^+(1)}$ is an instance
of the form $x_{i^+(j^+)}$, (ii) the contrastive example
for $x_{i^-(1)}$ is an instance of the form $x_{i^-(j^-)}$,
and (iii) $i^-(j^-) > i^+(j^+)$, then (I5) applies.
\end{enumerate}
Hence each possible response of the oracle leads to the exact
identification of the target concept.
\hfill$\Box$

\lemSubclass*
\iffalse
\begin{restatable}{lemma}{lemSubclass} \label{lem:min-dist-to-subC-MQ}
Suppose that $\cC$ is a concept class over $B_m$, and
$\cC'$ is a subclass of $\cC$ with the properties \\
\textbf{(P1)}
For each $\vec{a} \in B_m$ with $a_m = y_m$ and each $C \in \cC'$,
we have that $C(\vec{a}) = 1$.\\
\textbf{(P2)}
For each $\vec{a}$ with $a_m= y'_m$ and $a_{m-1}=y_{m-1}$ and each $C \in \cC'$,
we have that $C(\vec{a}) = 0$.\\
% In other words, each Boolean vector ending $1$ is a positive example for each $C \in \cC'$ while each Boolean vector ending $10$ is a negative example for each $C \in \cC'$.
Here $y_{m-1}, y_m \in \{0, 1\}$ are fixed, $y'_m = 1 - y_m$, $y'_{m-1} = 1 - y_{m-1}$, and $\cC''$ is the class of all concepts of the form
% ~(\ref{eq:hard-subclass}).
\begin{equation*} %\label{eq:hard-subclass}
%\cC'' = \{
(a_1,\ldots,a_{m-2}) \mapsto C(a_1,\ldots,a_{m-2}, y'_{m-1}, y'_{m})
\mbox{ for } C \in \cC'
%\}
\end{equation*}
Then $\cS_{\CS^d_{\min}}(\cC) \ge\cS_{\CS^d_{\min}}(\cC') \geq \cS[\MQ](\cC'')$, where $d$ is the Hamming distance.
%Then the number of contrast-oracle queries needed to learn $\cC$ (or $\cC'$) in the minimum-distance model is lower-bounded by the number of membership queries needed to learn $\cC''$.
\end{restatable}
\fi

\emph{Proof.}
Suppose that $L$ learns $\cC'$ from a contrast oracle in the minimum distance
model at the expense of $q$ queries. It suffices to show that $L$ can be
transformed into $L'$ which learns $\cC''$ from a membership oracle at
the expense of at most $q$ queries. To this end, let $C''$ be the target
concept in $\cC''$ and let $C'$ with $C'(\vec{a}y'_{m-1}y'_{m}) = C''(\vec{a})$ be
the corresponding concept in $\cC'$. Note that $L$ has not uniquely
identified $C'$ in $\cC'$ as long as the subfunction $C''$ is not uniquely
identified in $\cC''$. $L'$ can therefore uniquely identify $C''$ in $\cC''$
by maintaining a simulation of $L$ until $L$ has uniquely identified $C'$
in $\cC$. In order to explain how $L'$ simulates the contrast oracle, we
proceed by case distinction:
\begin{itemize}
\item
If $L$ chooses a query point of the form $\vec{a}y'_{m-1}y'_m$, then $L'$ chooses
the query point $\vec{a}$ and receives the label $b := C''(\vec{a})$
from its membership oracle. Then $L'$ returns $b$ and the point $\vec{a'}$
to $L$ where $\vec{a'} =\vec{a}y'_{m-1}y_m$ if $b=0$ and $\vec{a'} = \vec{a}y_{m-1}y'_m$
if $b=1$. Note that $(b,\vec{a'})$ is among the admissible answers of
the contrast oracle.
\item
If $L$ chooses a query point of the form $\vec{a}y_{m-1}y'_m$, then $L'$
returns the label $0$ and the additional point $\vec{a}y_{m-1}y_m$. Again, this is
among the admissible answers of the contrast oracle.
\item
If $L$ chooses a query point of the form $\vec{a}y_{m-1}y_m$, then $L'$
returns the label $1$ and the additional point $\vec{a}y_{m-1}y'_m$. Again,
this is among the admissible answers of the contrast oracle.
\item
If $L$ chooses a query-point of the form $\vec{a}y'_{m-1}y_m$, then $L'$
chooses the query point $\vec{a}$ and receives the label $b = C''(\vec{a})$
from its membership oracle. Then $L'$ returns the label $1$ and
the point $\vec{a'}$ to $L$ where $\vec{a'} = \vec{a}y'_{m-1}y'_m$ if $b=0$
and $\vec{a'} = \vec{a}y_{m-1}y'_m$ if $b=1$. Again, this is among the admissible
answers of the contrast oracle. Here note that, in case $b=1$, there is no admissible answer at distance 1 from the query point, so that a contrastive example at distance 2 can be chosen.
\end{itemize}
Clearly $L''$ can maintain this simulation until it reaches exact identification
of the target concept. Moreover, the number of query point chosen by $L'$
does not exceed the number of query points chosen by $L$.
\hfill$\Box$

\cordecisionlistPP*
%\setcounter{theorem}{15}
%\begin{restatable}{corollary}{cordecisionlistPP}\label{cor2:decision-list-lb}
%$\cS_{\CS^d_{\min}}(\DL^2_m)\ge 2^{m-2}-1$, where $d$ is the Hamming distance.
%At least $2^{m-2}-1$ contrast-oracle queries are needed for learning DL$_m^2$ in the minimum-distance model.
%\end{restatable}
\emph{Proof.}
Let $\cC = \mbox{DL}^2_m$. Moreover, let $\cC' \subseteq \cC$ be a subset of DL$^2_m$ of decision lists of the form
\[
\cL = [(v_m,1) , (v_{m-1},0) ,
\underbrace{(\ell_1,b_1) , \ldots (\ell_z,b_z) , b_{z+1}}_{= \cL' \in \mbox{\small DL}^1_{m-2}}] \in \mbox{DL}^2_m
\enspace ,
\]
where $z \ge 0$, $\ell_i \in \{v_1,\bar v_1,\ldots,v_{m-2},\bar v_{m-2}\}$
and $b_i \in \{0,1\}$. $\cC'$ satisfies properties (P1) and (P2) with $y_{m-1} = y_{m} = 1$ in Lemma~\ref{lem:min-dist-to-subC-MQ}. Then $\cC'' = \mbox{DL}^1_{m-2}$.  $\MONOMIALS^{m-2}$
is a subclass of DL$^1_{m-2}$ and $2^{m-2}-1$ membership queries are needed
for learning $\MONOMIALS^{m-2}$ \cite{KobayashiS09}.
\hfill$\Box$

%\setcounter{theorem}{16}

%\begin{restatable}{theorem}{
\thmsddl*
%} \label{thm:DL-SD}
%There exists a self-directed learner for $\DL_m$ that makes, on any  target list $C^*\in\DL_m$, at most $4km$ mistakes where $k$ is the number of blocks in  $C^*$.
%\end{restatable}
%\thmsddl*
\emph{Proof.}
Recall $B_m = \{0,1\}^m$. For any set of literals $\ell_{i_1}$, \dots, $\ell_{i_j}$, we use $B_m(\ell_{i_1} = b_1,\ldots,\ell_{i_j} = b_j)$ to denote the $(m-j)$-dimensional subcube resulting from fixing the value of $\ell_{i_t}$ to $b_t$, for $t\in[j]$. We call a literal $\ell$ $b$-pure if the target 
list $C^*$ assigns label $b$ to every point in $B_m(\ell=1)$.
 
The $b$-pureness of $\ell = v_i$ (resp.\ of $\ell = \bar v_i$) can be checked
by testing whether $C^*$ restricted to $B_m(v_i=1)$  (resp.~to $B_m(v_i=0)$)
degenerates to the constant-$b$ function. This is done at the expense
of at most $2$ mistakes per literal: The learner predicts 0 for a first point in $B_m(\ell=1)$, receives the correct label $b$ (which may or may not count as one mistake), and then predicts $b$ for further points in $B_m(\ell=1)$ until a mistake is made. If no mistake is made on further points in $B_m(\ell=1)$, then $\ell$ is $b$-pure; otherwise it is not pure. Hence, after having made at most $4m$ mistakes,
the learner $L$ knows all pure-literals, say $\ell_1,\ldots,\ell_t$,
and $L$ also knows the (unique) bit $b \in \{0,1\}$ with respect to
which $\ell_1,\ldots,\ell_t$ are pure.
 
At this point $L$ knows the first block $(\ell_1,b), \ldots (\ell_t,b)$
of the list $C^*$. $L$ can now proceed iteratively in order to learn the
remaining blocks. As for the second block, all query points are taken
from the $(m-t)$-dimensional subcube $B_m(\ell_1 = 1-b,\ldots,\ell_t = 1-b)$
so that they are not absorbed by the first block. In this way, $C^*$ is identified
blockwise at the expense of at most $4m$ queries per block.
\hfill$\Box$

\thmmdnf*
\iffalse
\setcounter{theorem}{20}
\begin{restatable}{theorem}{thmmdnf}\label{thm:mdnf2}
   %Then \YC{drop ``Then''?} 
   $\cS_{CS^{(d_{\cC_t})}_{\min}}(\MDNF{m}{s}{z}) \leq s = \operatorname{SD}(\MDNF{m}{s}{z})$, where $\cC_t=(\MDNF{m}{s}{z})_t$ for any $t$.
\end{restatable} 
\fi
The proof of this theorem makes use of the following terminology. Suppose $\vec a=(a_1,\ldots,a_m),\vec b=(b_1,\ldots,b_m)\in B^m$ for some $m$. We write $\vec{a} \leq \vec{b}$ if $a_i=1$ implies $b_i=1$. Moreover, we implicitly identify the Boolean vector $\vec a$ with the corresponding monotone monomial.

First we introduce a helpful lemma. 

\begin{lemma}\label{lem:mon-MDNF}
    Let $\vec a, \vec 
    b, \vec c \in \{0, 1\}^m$ such that $\vec a \le\vec b\le\vec c$. Then for any $\cC' \subseteq \MDNF{m}{s}{z}$ we have $d_{\cC'}(\vec a, \vec b) \leq d_{\cC'}(\vec a, \vec c)$.
\end{lemma}
\emph{Proof.}
    Consider any $C \in \MDNF{m}{s}{z}$. Since $C$ is monotone,  if $C(\vec a) = 1$ then we have $C(\vec b) = 1, C(\vec c) = 1$. Similarly, if $C(\vec a) = 0$ and $C(\vec b) = 1$ then $C(\vec c) = 1$ as well. This completes the proof. 
\hfill$\Box$

\emph{Proof.} \emph{(of Theorem~\ref{thm:mdnf2}).}
The claim $\operatorname{SD}(\MDNF{m}{s}{z})=s$ was proven by \cite{GoldmanS94} and already stated in Theorem~19. It suffices to define a learner that witnesses $\cS_{CS^{(d_{\cC_t})}_{min}}(\MDNF{m}{s}{z}) \leq s$.

    Define a learner that asks, in each of $s$ iterations, a query for the vector $\vec 0$. Since $\vec 0$ has label 0, all the contrastive examples have label 1. Let $\vec{b}^t$ be the contrastive example for the $t$th query.

   First, we argue that $\vec{b}^1$ is a monomial in the target concept $C^*$. Suppose $\vec{b}^1$ is not a monomial. Since it has label 1, there is a monomial $\vec{b'}\le\vec{b}^1$. Using Lemma~\ref{lem:mon-MDNF}, we have $d_{\cC_1}(0, \vec{b'}) \leq d_{\cC_1}(0, \vec{b}^1)$. Moreover, since $\vec{b}^1 \neq \vec{b'}$, there exists $C \in \MDNF{m}{s}{z}$ such that $C(\vec{b'}) = 0$, but $C(\vec{b}^1) = 1$. Thus, $d_{\cC_1}(0, \vec{b'}) < d_{\cC_1}(0, \vec{b}^1)$. Therefore $\vec{b'}$ should have been the contrastive example, which is a contradiction.

  Next, we prove inductively that $\{\vec{b}^1, ..., \vec{b}^s\}$ is the set of monomials in $C^*$. %We already proved the base of induction in last paragraph. 
  Let $\{\vec{b}^1, ..., \vec{b}^{t-1}\}$ be $t - 1$ distinct monomials in $C^*$. We prove that $\vec{b}^t$ is a monomial in $C^*$ distinct from $\vec{b}^1, ..., \vec{b}^{t-1}$.

  Note that $\cC_{t-1}$ is the set of concepts that have label 1 on all of $\vec{b}^1, ..., \vec{b}^{t-1}$. We also know that all concepts give label 0 to the all zero vector. Thus, according to Lemma~\ref{lem:mon-MDNF} for any $\vec a$ with $\vec{b}^q\le \vec a$ for some $q \in [t-1]$, we have $d_{\cC_1}(0, \vec a) \geq d_{\cC_1}(0, \vec{b}^q) = 1$ (maximum distance). Thus, $\vec{b}^q\not\leq\vec{b}^t$. With an argument similar to one we used for the base case, $\vec{b}^t$ is also a monomial. This completes the proof.
\hfill$\Box$

%\section*{Ethical Statement}

%There are no ethical issues.

%\section*{Acknowledgments}

\end{document}